\documentclass[10pt,twocolumn,twoside]{IEEEtran}
\usepackage{amsmath,amsopn,amssymb,stackrel}
\usepackage{graphicx,xspace,color,soul}
\usepackage{epsfig,subfigure}
\usepackage{longtable,multirow}
\usepackage{array,float}
\usepackage{cite,citesort}
\usepackage{url}
\renewcommand{\vec}[1]{\mathbf{#1}}
\usepackage{algorithm,algorithmicx,algpseudocode}

\newcommand\ceil[1]{\lceil#1\rceil}

\usepackage{epstopdf}
\usepackage{stfloats}

\newcounter{mytempeqncnt}
\graphicspath{{figs/}}

\begin{document}
\title{Joint Denoising / Compression of Image Contours via Shape Prior and Context Tree}

\author{
Amin Zheng~\IEEEmembership{Student Member,~IEEE},
Gene Cheung~\IEEEmembership{Senior Member,~IEEE},
Dinei Florencio~\IEEEmembership{Fellow,~IEEE}
\begin{small}
\thanks{A. Zheng is with 
        Department of Electronic and Computer Engineering, 
        The Hong Kong University of Science and Technology, 
        Clear Water Bay, Hong Kong, China
        (e-mail: amzheng@connect.ust.hk).}
        \thanks{G. Cheung is with 
        National Institute of Informatics, 2-1-2, Hitotsubashi, Chiyoda-ku,
        Tokyo, Japan 101--8430 
        (e-mail: cheung@nii.ac.jp).}
        \thanks{D. Florencio is with  
        Microsoft Research,
        Redmond, WA USA 
        (e-mail: dinei@microsoft.com).}
\end{small}
}%
\maketitle
\vspace{0.1in}

\begin{abstract}
With the advent of depth sensing technologies, the extraction of object contours in images---a common and important pre-processing step for later higher-level computer vision tasks like object detection and human action recognition---has become easier. 
However, acquisition noise in captured depth images means that detected contours suffer from unavoidable errors. 
In this paper, we propose to jointly denoise and compress detected contours in an image for bandwidth-constrained transmission to a client, who can then carry out aforementioned application-specific tasks using the decoded contours as input.
We first prove theoretically that in general a joint denoising / compression approach can outperform a separate two-stage approach that first denoises then encodes contours lossily.
Adopting a joint approach, we first propose a burst error model that models typical errors encountered in an observed string $\vec{y}$ of directional edges.
We then formulate a rate-constrained maximum a posteriori (MAP) problem that trades off the posterior probability $P(\hat{\vec{x}} | \vec{y})$ of an estimated string $\hat{\vec{x}}$ given $\vec{y}$ with its code rate $R(\hat{\vec{x}})$. 
We design a dynamic programming (DP) algorithm that solves the posed problem optimally, and propose a compact context representation called total suffix tree (TST) that can reduce complexity of the algorithm dramatically.
Experimental results show that our joint denoising / compression scheme outperformed a competing separate scheme in rate-distortion performance noticeably.
\end{abstract}


\IEEEpeerreviewmaketitle

\section{Introduction}
\label{sec:intro}
Advances in depth sensing technologies like Microsoft Kinect 2.0 means that depth images---per pixel distances between physical objects in a 3D scene and the camera---can now be captured easily and inexpensively.
Depth imaging has in turn eased the extraction of object contours in a captured image, which was traditionally a challenging computer vision problem \cite{grigorescu2003contour}.
Detected contours can be used to facilitate recent advanced image coding schemes; if object contours are compressed efficiently as side information (SI), they can enable new edge-adaptive techniques such as graph Fourier transform (GFT) coding \cite{hu15,hu15spl} and motion prediction of arbitrarily shaped blocks \cite{daribo14}. 
Further, coded object contours can also be transmitted to a central cloud for computationally expensive application-specific tasks such as object detection or human action recognition \cite{weinland2011survey}, at a much lower coding cost than the original captured depth video.

Unfortunately, captured depth images are often corrupted by acquisition noise, and hence unavoidably the detected contours also contain errors. 
In this paper, we propose to jointly denoise and compress detected object contours in images.
First, we prove theoretically that in general a joint denoising / compression approach outperforms a two-stage separate approach that first denoises an observed contour then compresses the denoised version lossily.
Adopting a joint approach, we first propose a burst error model that captures unique characteristics of typical errors encountered in detected contours.
We then formulate a rate-constrained \textit{maximum a posteriori} (MAP) problem that trades off the posterior probability $P(\hat{\vec{x}} | \vec{y})$ of an estimated contour $\hat{\vec{x}}$ given observed contour $\vec{y}$ with its code rate $R(\hat{\vec{x}})$. 
Given our burst error model, we show that the negative log of the likelihood $P(\vec{y}|\vec{x})$ can be written intuitively as a simple sum of burst error events, error symbols and burst lengths. 
Further, we construct a geometric prior $P(\vec{x})$ stating intuitively that contours are more likely straight than curvy. 

We design a dynamic programming (DP) \cite{dreyfus1977art} algorithm that solves the posed problem optimally, and propose a compact context representation called \textit{total suffix tree} (TST) that can reduce the algorithm complexity dramatically.
Experimental results show that our joint denoising / compression scheme outperformed a competing separate scheme in RD performance noticeably.
We note that, to the best of our knowledge, we are the first in the literature\footnote{An early version of this work appeared as a conference paper \cite{zheng16}.} to formally address the problem of joint denoising / compression of detected image contours.

The outline of the paper is as follows. 
We first overview related works in Section\;\ref{sec:related}. 
We pose our rate-constrained MAP problem in Section \ref{sec:problem}, and define the corresponding error and rate terms in Section\;\ref{sec:error} and \ref{sec:rate}, respectively.
We describe our proposed optimal algorithm in Section\;\ref{sec:algorithm}. 
Experimental results are presented in Section\;\ref{sec:results}, and we conclude in Section\;\ref{sec:conclude}. 

\section{Related Work}
\label{sec:related}
\subsection{Contour Coding}
\label{subsec:related_contour_coding}

\subsubsection{Lossless Contour Coding}
\label{subsubsec:related_lossless}

Most works in lossless contour coding \cite{daribo14,freeman1978application,liu2005efficient,DCC1st1991,chan1995highly,estes1995efficient,turner1996efficient,egger1996region,jordan1998shape} first convert an image contour into a \textit{chain code} \cite{freeman1961}: a sequence of symbols each representing one of four or eight possible \textit{absolute} directions on the pixel grid. 
Alternatively, a \textit{differential chain code} (DCC) \cite{freeman1978application} that specifies \textit{relative} directions instead can be used.
DCC symbols are entropy-coded using either Huffman \cite{liu2005efficient} or arithmetic coding \cite{DCC1st1991} given symbol probabilities. 
The challenge is to estimate conditional probabilities for DCC symbols given a set of training data; this is the main problem we discuss in Section \ref{sec:rate}. 

\cite{daribo12icip,daribo14} propose a linear geometric model to estimate conditional probabilities of the next DCC symbol. 
In summary, given a window of previous edges, a line-of-best-fit that minimizes the sum of distances to the edges' endpoints is first constructed. Then the probability of a candidate direction for the next symbol is assumed inversely proportional to the angle difference between the direction and the fitted line. 
This scheme is inferior in estimating symbol probabilities compared to context models, because there are only a few possible angle differences for a small number of previous edges, limiting the expressiveness of the model.

An alternative approach is \textit{context modeling}: given a window of $l$ previous symbols (context) $\mathbf{x}^{i-1}_{i-l}$, compute the conditional probability $P(x_{i} | \mathbf{x}^{i-1}_{i-l})$ of the next symbol $x_{i}$ by counting the number of occurrences of $\mathbf{x}^{i-1}_{i-l}$ followed by $x_{i}$ in the training data. 
In \cite{kaneko1985encoding,DCC1st1991,chan1995highly,estes1995efficient,turner1996efficient}, Markov models of fixed order up to eight are used for lossless coding. 
However, in applications where the training data is limited, there may not be enough occurrences of $\mathbf{x}^{i-1}_{i-l}$ to reliably estimate the conditional probabilities. 

\textit{Variable-length Context Tree} (VCT) \cite{rissanen1983universal,begleiter2004prediction} provides a more flexible approach for Markov context modeling by allowing each context to be of variable length. 
There are many ways to construct a VCT; \textit{e.g.}, Lempel-Ziv-78 (LZ78) \cite{ziv1977universal} and prediction by partial matching (PPM) \cite{cleary1984data,moffat1990implementing}.
LZ78 constructs a dictionary from scratch using the input data directly as training data. 
The probability estimation quality varies depending on the order of first appearing input symbols.
PPM considers all contexts restricted by a maximum length with non-zero occurrences in the training data when building a VCT.
PPM has an efficient way to deal with the zero frequency problem \cite{begleiter2004prediction}, where the input symbol along with the context does not occur in the training data, by reducing the context length. 
PPM is widely used for lossless contour coding \cite{egger1996region,jordan1998shape} because of its efficiency.
In this paper, we use PPM for contour coding, as described in Section \ref{sec:rate}.

\subsubsection{Lossy Contour Coding}
\label{subsubsec:related_lossy}

Lossy contour coding approaches can be classified into two categories: DCC-based \cite{zahir2007new,yuan2015contour} and vertex-based \cite{katsaggelos1998mpeg,sohel2007new,lai2010arbitrary} approaches. 
In \cite{zahir2007new,yuan2015contour}, the DCC strings are approximated by using line smoothing techniques. 
In contrast, vertex-based approaches select representative key points along the contour for coding at the encoder and interpolation at the decoder.
Because vertex-based approaches are not suitable for lossless contour coding and we consider joint denoising / compression of contours for a wide range of bitrates, 
we choose a DCC-based approach and compute the optimal rate-constrained MAP contour estimate for coding using PPM.

\subsection{Contour Denoising}
\label{subsec:related_contour_denoising}
Contour noise removal is considered in \cite{yu1997efficient,daribo14,hoang2011edge,reisert2008equivariant,lu2015rapid,zhong2010convergence}.
In \cite{yu1997efficient}, the noisy pixels along the chain code are first removed, then the processed chain code is decomposed into a set of connected straight lines.
In \cite{daribo14}, a chain code is simplified by removing ``irregularities", which are predefined non-smooth edge patterns.
Both \cite{yu1997efficient} and \cite{daribo14} are contour smoothing techniques, which does not guarantee the similarity between the original contour and the smoothed contour.
Sparse representation and non-linear filter are used in \cite{hoang2011edge} and \cite{reisert2008equivariant} respectively to enhance the contours in a binary document image, with the objective of removing the noise in the binary image.
\cite{lu2015rapid,zhong2010convergence} denoise the image contours using Gaussian filter by averaging the coordinates of the points on the contours.
Gaussian filter is widely used for removing noise from signal, but the Gaussian model does not specifically reflect the statistics of the noise on the image contours.
In contrast, we propose a burst error model to describe the errors on the image contours, as discussed in Section \ref{sec:error}.

Considering the problem of both denoising and compression of the image contours, the denoised contour by \cite{yu1997efficient,daribo14,hoang2011edge,reisert2008equivariant,lu2015rapid,zhong2010convergence} may require a large encoding overhead.
We perform a MAP estimate to denoise an observed contour subject to a rate constraint. We will show in Section \ref{sec:results} that we outperform a separate denoising / compression approach.

\section{Problem Formulation}
\label{sec:problem}
\begin{figure}[t]

\begin{minipage}[b]{1\linewidth}
  \centering
  \centerline{\includegraphics[width=4cm]{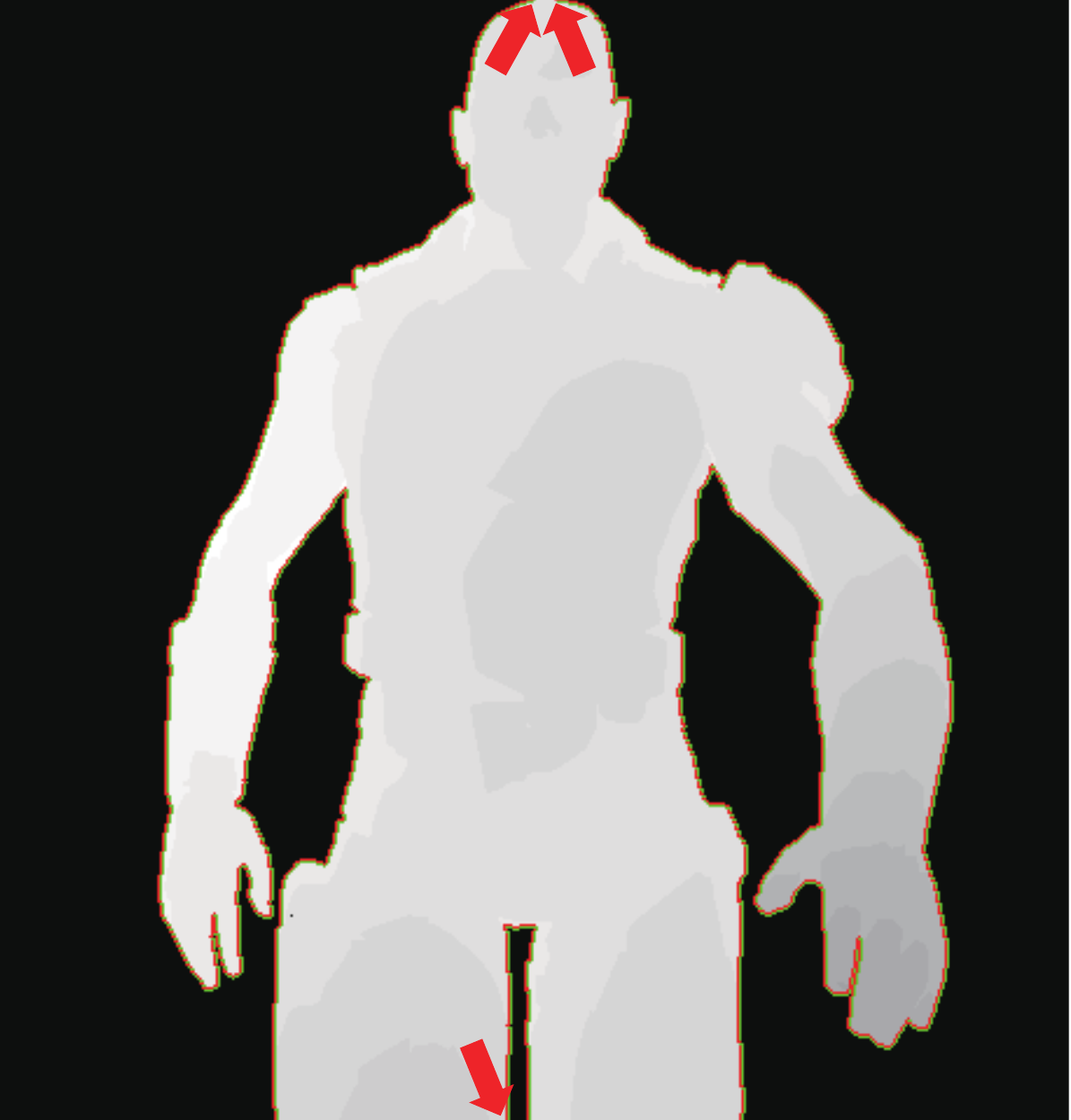}}
\end{minipage}

\vspace{-0.3cm}
\caption{Depth image with three detected contours.
The detected contours are the edges between the green and the red pixels. 
Initial points of the contours are indicted by red arrows.}
\label{fig:contour_definition}
\end{figure}

We first describe our contour denoising / compression problem at a high level.
We assume that one or more object contours in a noise-corrupted image have first been detected, for example, using a method like gradient-based edge detection \cite{daribo14}.
Each contour is defined by an initial point and a following sequence of connected ``between-pixel" edges on a 2D grid that divide pixels in a local neighbourhood into two sides.
As an example, three contours in one video frame of \texttt{Dude} are drawn in Fig.\;\ref{fig:contour_definition}.
For each detected (and noise-corrupted) contour, the problem is to estimate a contour that is highly probable \textit{and} requires few bits for encoding.

We discuss two general approaches for this problem---joint approach and separate approach---and compare and contrast the two.
Adopting the joint approach, we formulate a rate-constrained MAP problem.

\begin{figure}[t]

\begin{minipage}[b]{.62\linewidth}
  \centering
  \centerline{\includegraphics[width=5cm]{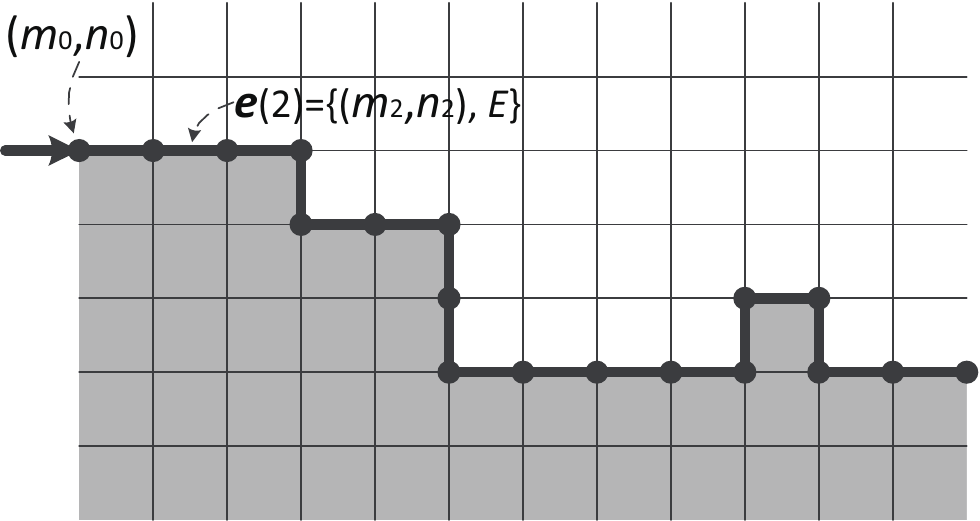}}
  \centerline{(a)}\medskip
\end{minipage}
\hfill
\begin{minipage}[b]{0.36\linewidth}
  \centering
  \centerline{\includegraphics[width=3cm]{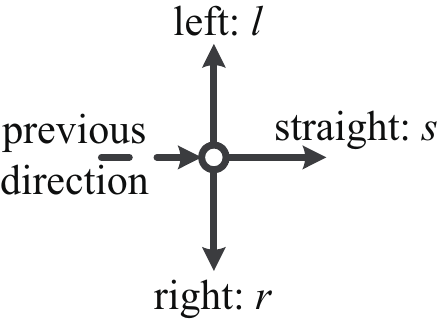}}
  \centerline{(b)}\medskip
\end{minipage}

\vspace{-0.2cm}
\caption{{(a) An example of an observed length-17 contour: $\!\texttt{E}\!-\!\texttt{s}\!-\!\texttt{s}\!-\!\texttt{r}-\!\texttt{l}\!-\!\texttt{s}\!-\!\texttt{r}-\!\texttt{s}\!-\!\texttt{l}\!-\!\texttt{s}\!-\!\texttt{s}\!-\!\texttt{s}\!-\!\texttt{l}\!-\!\texttt{r}\!-\!\texttt{r}\!-\!\texttt{l}\!-\!\texttt{s}$. $(m_0, n_0)$ is the coordinate of the initial point and $\vec{e}(2)=\{(m_0+2,n_0),E\}$. (b) Three relative directions.}}
\label{fig:DCC}
\end{figure}

\subsection{Joint and Separate Approaches}
\label{subsec:analysis_jointSeparate}

\subsubsection{Definitions}

We first convert each contour into a \textit{differential chain code} (DCC)\cite{freeman1978application}.
A length-$L_{\vec{z}}$ contour $\vec{z}$ can be compactly described as a symbol string denoted by $[z_1,z_2,\cdots,z_{L_{\vec{z}}}]$.
As shown in Fig.\;\ref{fig:DCC}, the first symbol $z_1$ is chosen from a size-four alphabet, $\mathcal{D}=\{\texttt{N},\texttt{E},\texttt{S},\texttt{W}\}$, specifying the four \textit{absolute} directions \texttt{north}, \texttt{east}, \texttt{south} and \texttt{west} with respect to a 2D-grid start point, \textit{i.e.}, $z_1\in \mathcal{D}$. 
Each subsequent DCC symbol $z_i$, $i\geq 2$, is chosen from a size-three alphabet, $\mathcal{A}=\{\texttt{l}, \texttt{s}, \texttt{r}\}$, specifying the three \textit{relative} directions \texttt{left}, \texttt{straight} and \texttt{right} with respect to the previous symbols $z_{i-1}$, \textit{i.e.}, $z_i\in \mathcal{A}$ for $i\geq 2$.
Denote by $\vec{z}^{j}_{i} = [{z}_{j},{z}_{j-1},\ldots,{z}_{i}]$, $i<j$ and $i,j \in \mathbb{Z}^+$, a \textit{sub-string} of length $j-i+1$ from the $i$-th symbol $z_i$ to the $j$-th symbol $z_j$ in reverse order. 

Alternatively, one can represent the $i$-th edge of contour $\vec{z}$ geometrically on the 2D grid as $\vec{e}_{\vec{z}}(i)=\{(m_i,n_i), d_i\}$, where $(m_i, n_i)$ is the 2D coordinate of the ending point of the edge and $d_i \in \mathcal{D}$ is the absolute direction, as shown in Fig.\;\ref{fig:DCC}(a).
This representation is used later when describing our optimization algorithm.

\subsubsection{Joint Problem Formulation}

Denote by $\vec{y} \in \mathcal{S}$ and $\vec{x} \in \mathcal{S}$ the observed and ground truth DCC strings respectively, where $\mathcal{S}$ is the space of all DCC strings of finite length. 
As done in \cite{sun13,sun14} for joint denoising / compression of multiview depth images\footnote{We note that though \cite{sun13,sun14} also formulated a rate-constrained MAP problem for the joint denoising / compresion problem, there was no theoretical analysis on why a joint approach is better in general.}, we follow a rate-constrained MAP formulation and define the objective as finding a string $\hat{\vec{x}} \in \mathcal{S}$ that maximizes the posterior probability $P(\hat{\vec{x}}|\vec{y})$, subject to a rate constraint on chosen $\hat{\vec{x}}$:
\begin{equation}
\label{eq:orig_objective}
\underset{\hat{\vec{x}}\in  \mathcal{S}}{\max}\ P(\hat{\vec{x}}|\vec{y}) ~~~ \mbox{s.t.} ~ 
R(\hat{\vec{x}})\leq  R_{\max}
\end{equation}
where $R(\hat{\vec{x}})$ is the bit count required to encode string $\hat{\vec{x}}$, and $R_{\max}$ is the bit budget. 

(\ref{eq:orig_objective}) essentially seeks one solution to \textit{both} the estimation problem \textit{and} the compression problem at the same time.
The estimated rate-constrained DCC string from (\ref{eq:orig_objective}) is then \textit{losslessly} compressed at the encoder.

\subsubsection{Separate Problem Formulation}

An alternative approach is to first solve an unconstrained MAP estimation sub-problem, then \textit{lossily} compress the estimated DCC string subject to the rate constraint as follows:
\begin{equation}
\vec{x}_{MAP} = \arg\,\underset{\hat{\vec{x}}\in  \mathcal{S}}{\max} \ P(\hat{\vec{x}}|\vec{y})
\label{eq:separate_estimation}
\end{equation} 
\begin{equation}
\underset{\hat{\vec{x}}\in  \mathcal{S}}{\max}\ P(\hat{\vec{x}}|\vec{x}_{MAP}) ~~~ \mbox{s.t.} ~ 
R(\hat{\vec{x}})\leq  R_{\max}
\label{eq:separate_compression}
\end{equation} 
where $\vec{x}_{MAP}$ is an optimal solution to the unconstrained estimation sub-problem (\ref{eq:separate_estimation}).
The lossy compression sub-problem (\ref{eq:separate_compression}) aims to find a DCC string $\vec{\hat{x}}$ which is closest to $\vec{x}_{MAP}$ in probability given bit budget $R_{\max}$.
In the sequel we call (\ref{eq:orig_objective}) the \textit{joint} approach, and (\ref{eq:separate_estimation}) and (\ref{eq:separate_compression}) the \textit{separate} approach.

Since the contour denoising / compression problem is inherently a joint estimation /  compression problem, the solution to the joint approach should not be worse than the solution to the separate approach.
The question is under what condition(s) the joint approach is strictly better than the separate approach.

\subsection{Comparison between Joint and Separate Approaches}
\label{subsec:analysis_comparison}

\subsubsection{Algebraic Comparison}

To understand the condition(s) under which the joint approach is strictly better than the separate approach, we first rewrite $P(\hat{\vec{x}}|\vec{x}_{MAP})$ in (\ref{eq:separate_compression}) using the law of total probability by introducing an observation $\hat{\vec{y}}$:
\begin{equation}
P(\hat{\vec{x}}|\vec{x}_{MAP}) = \sum_{\hat{\vec{y}}} P(\hat{\vec{x}}|\vec{x}_{MAP},\hat{\vec{y}})\,P(\hat{\vec{y}}|\vec{x}_{MAP}).
\label{eq:comparison_total}
\end{equation}
$P(\hat{\vec{y}}|\vec{x}_{MAP})$ can be further rewritten using Bayes' rule:
\begin{equation}
P(\hat{\vec{y}}|\vec{x}_{MAP}) = \frac{P(\vec{x}_{MAP}|\hat{\vec{y}})P(\hat{\vec{y}})}{P(\vec{x}_{MAP})}.
\label{eq:comparison_bayes}
\end{equation}
From (\ref{eq:separate_estimation}), $P(\vec{x}_{MAP}|\hat{\vec{y}}) = 1$ if $\vec{x}_{MAP}$ is the MAP solution given this observation $\hat{\vec{y}}$; otherwise, $P(\vec{x}_{MAP}|\hat{\vec{y}}) = 0$:
\begin{equation}
P(\vec{x}_{MAP}|\hat{\vec{y}}) = 
\begin{cases}
1, & \text{if } \vec{x}_{MAP} = \arg\,\underset{\tilde{\vec{x}}\in  \mathcal{S}}{\max} \ P(\tilde{\vec{x}}|\hat{\vec{y}}) \\
0, & \text{otherwise}
\end{cases}.
\label{eq:comparison_cases}
\end{equation}

Combining (\ref{eq:separate_estimation}) and (\ref{eq:comparison_total}) to (\ref{eq:comparison_cases}), $P(\hat{\vec{x}}|\vec{x}_{MAP})$ is rewritten as follows:
\begin{equation}
\begin{split}
&\sum_{\hat{\vec{y}}} P(\hat{\vec{x}}|\vec{x}_{MAP},\hat{\vec{y}}) P(\vec{x}_{MAP}|\hat{\vec{y}}) P(\hat{\vec{y}}) \\
&= \sum_{\hat{\vec{y}}|\vec{x}_{MAP} = \arg\,\underset{\tilde{\vec{x}}\in  \mathcal{S}}{\max} \ P(\tilde{\vec{x}}|\hat{\vec{y}})} P(\hat{\vec{x}}|\vec{x}_{MAP},\hat{\vec{y}}) P(\hat{\vec{y}}) \\
&= \sum_{\hat{\vec{y}}|\vec{x}_{MAP} = \arg\,\underset{\tilde{\vec{x}}\in  \mathcal{S}}{\max} \ P(\tilde{\vec{x}}|\hat{\vec{y}})} P(\hat{\vec{x}}|\hat{\vec{y}}) P(\hat{\vec{y}})
\end{split}
\end{equation}
where $P(\hat{\vec{x}}|\vec{x}_{MAP},\hat{\vec{y}}) = P(\hat{\vec{x}}|\hat{\vec{y}})$, since $\vec{x}_{MAP}$ is a function of $\hat{\vec{y}}$, \textit{i.e.}, $\vec{x}_{MAP} = \arg\,\underset{\tilde{\vec{x}}\in  \mathcal{S}}{\max} \ P(\tilde{\vec{x}}|\hat{\vec{y}})$. 
We ignore $P(\vec{x}_{MAP})$ in (\ref{eq:comparison_bayes}), because $\vec{x}_{MAP}$ (and hence $P(\vec{x}_{MAP})$) is fixed and does not affect the maximization.

Thus, we can rewrite the separate approach as:
\begin{equation}
\underset{\hat{\vec{x}}\in  \mathcal{S}}{\max}\ \sum_{\hat{\vec{y}}|\vec{x}_{MAP} = \arg\,\underset{\tilde{\vec{x}}\in  \mathcal{S}}{\max} \ P(\tilde{\vec{x}}|\hat{\vec{y}})} P(\hat{\vec{x}}|\hat{\vec{y}}) P(\hat{\vec{y}}) ~~~ \mbox{s.t.} ~ 
R(\hat{\vec{x}})\leq  R_{\max}.
\label{eq:comprison_separate}
\end{equation}

Comparing (\ref{eq:comprison_separate}) to the joint approach (\ref{eq:orig_objective}), we can conclude the following: 
if there is only one observation $\hat{\vec{y}}$ that corresponds to $\vec{x}_{MAP}$, \textit{i.e.}, $\vec{x}_{MAP} = \arg\,\underset{\tilde{\vec{x}}\in  \mathcal{S}}{\max} \ P(\tilde{\vec{x}}|\hat{\vec{y}})$, then the solutions to the separate and joint approaches are the same---they both maximize $P(\hat{\vec{x}}|\vec{y})$ subject to $R(\hat{\vec{x}}) \leq R_{\max}$.
This makes intuitive sense: if there is a one-to-one correspondence between $\vec{x}_{MAP}$ and observation $\vec{y}$, then knowing $\vec{x}_{MAP}$ is as good as knowing $\vec{y}$.
On the other hand, if there are more than one observation $\hat{\vec{y}}$ corresponding to $\vec{x}_{MAP}$, \textit{i.e.}, there are  more than one term in the sum in (\ref{eq:comprison_separate}), then the solutions between the two approaches can be far apart.

\begin{figure}[t]

\begin{minipage}[b]{.48\linewidth}
  \centering
  \centerline{\includegraphics[width=4.5cm]{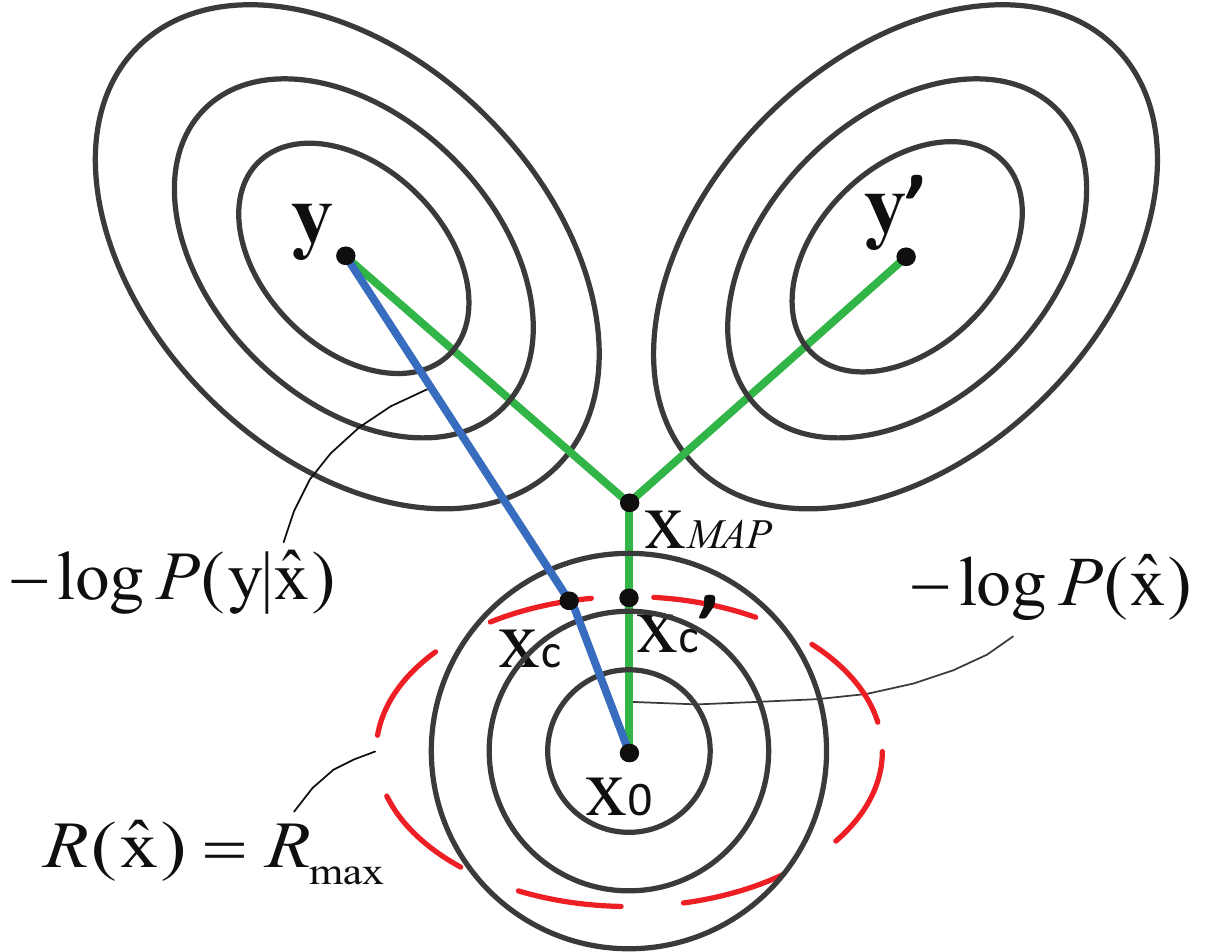}}
  \centerline{(a)}\medskip
\end{minipage}
\hfill
\begin{minipage}[b]{0.48\linewidth}
  \centering
  \centerline{\includegraphics[width=2.7cm]{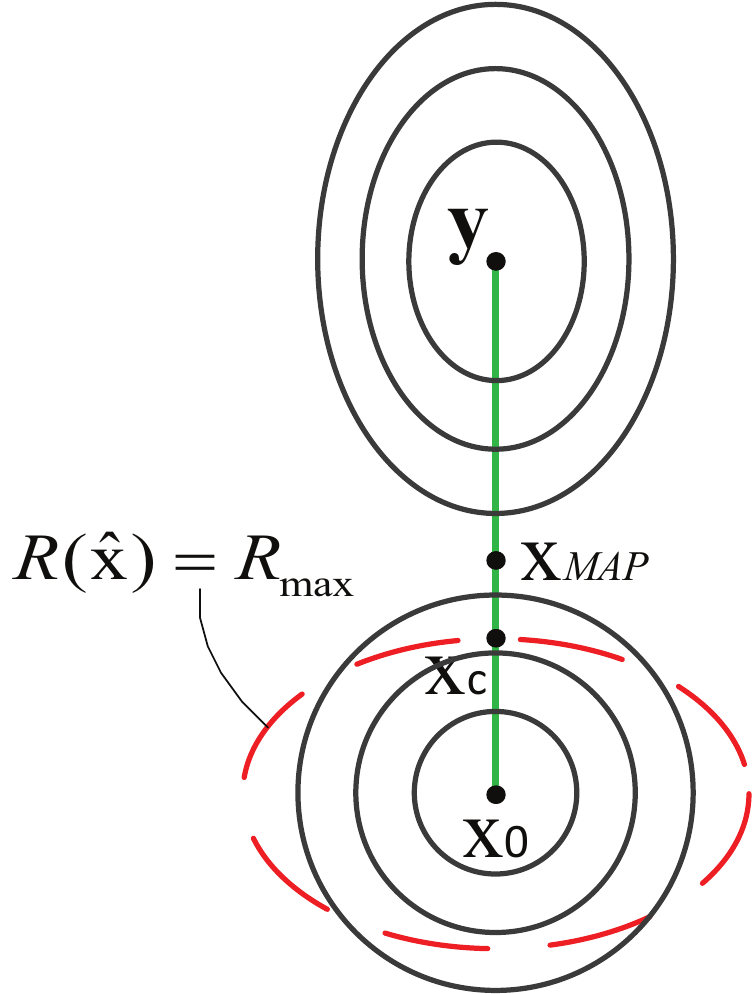}}
  \centerline{(b)}\medskip
\end{minipage}

\vspace{-0.3cm}
\caption{(a) Scenario where the joint approach is better than the separate approach.
(b) Scenario where the joint approach is same as the separate approach.}
\label{fig:joint_separate}
\end{figure}

\subsubsection{Geometric Comparison}

To develop intuition on the difference between the joint and separate approaches, we examine the problem from a geometric viewpoint by representing each DCC string as a point in a high-dimensional space.
By Bayes' rule, maximization of the posterior $P(\hat{\vec{x}}|\vec{y})$ is the same as maximization of the product of the likelihood $P(\vec{y}|\hat{\vec{x}})$ and the prior $P(\hat{\vec{x}})$, or minimization of $-\log P(\vec{y}|\hat{\vec{x}}) -\log P(\hat{\vec{x}})$.
As shown in Fig.\;\ref{fig:joint_separate}, each point in the space denotes a DCC string\footnote{We can map a DCC string into a point in a $N$-dimensional space as follows. First divide the string on the 2D grid into $(N-1)/2$ segments of equal length. The $x$- and $y$-coordinates of the end point of each segment on the 2D grid becomes an index for each segment. The resulted $N-1$ index numbers, plus a number to denote the length of each segment, becomes an $N$-dimension coordinate in an $N$-dimensional space.}.
Specifically, $\vec{y}$ and $\vec{y}^{\prime}$ in Fig.\;\ref{fig:joint_separate}(a) denote two different observations that would yield the same unconstrained MAP solution $\vec{x}_{MAP}$, \textit{i.e.}, $\vec{x}_{MAP} = \arg \max_{\hat{\vec{x}}} P(\hat{\vec{x}}|\vec{y}) = \arg \max_{\hat{\vec{x}}} P(\hat{\vec{x}}|\vec{y}')$. 

Denote by $\vec{x}_0$ the DCC string with the largest prior $P(\vec{x}_0)$, \textit{e.g.}, the string corresponding to the ``straightest" contour.
For illustration, we draw \textit{isolines} centered at $\vec{y}$ (or $\vec{y}^{\prime}$), each denoting the set of strings that have the same metric distance $D_l(\vec{y},\vec{x})$---negative log likelihood $-\log P(\vec{y}|\hat{\vec{x}})$---given observation $\vec{y}$ (or $\vec{y}^{\prime}$).
Similarly, we draw isolines centered at $\vec{x}_0$ to denote strings with the same metric distance $D_p(\vec{x}_0,\vec{x})$---negative log prior $-\log P(\hat{\vec{x}})$. 
Thus, maximizing $P(\hat{\vec{x}}|\vec{y})$ is equivalent to identifying a point $\vec{x}$ in the $N$-dimensional space that minimizes the sum of distance $D_l(\vec{y},\vec{x})$ and $D_p(\vec{x}_0,\vec{x})$.

The joint approach is then equivalent to finding a point $\hat{\vec{x}}$ inside the isoline defined by the rate constraint $R(\vec{\hat{x}})=R_{\max}$---\textit{feasible rate region}---to minimize the sum of the two aforementioned distances.
On the other hand, the separate approach is equivalent to first finding the point $\vec{x}_{MAP}$ to minimize $D_l(\vec{y},\vec{x}_{MAP}) + D_p(\vec{x}_0,\vec{x}_{MAP})$, then find a point $\hat{\vec{x}}$ inside the feasible rate region to minimize $D_l(\vec{x}_{MAP}, \hat{\vec{x}})$, 
\textit{i.e.}, the distance from $\hat{\vec{x}}$ to $\vec{x}_{MAP}$.

In Fig.\;\ref{fig:joint_separate}(a), $\vec{x}_c^{\prime}$ is the solution to the separate approach, which is the point bounded by the rate constraint with minimum distance $D_l(\vec{x}_{MAP}, \vec{x}_c^{\prime})$ to $\vec{x}_{MAP}$.
Note that the solution to the joint approach denoted by $\vec{x}_c$ may be different from $\vec{x}_c^{\prime}$; \textit{i.e.}, the sum of the two distances $D_l(\vec{y},\vec{x}_c) + D_p(\vec{x}_0,\vec{x}_c)$ from $\vec{x}_c$ is smaller than the distance sum from $\vec{x}_c^{\prime}$. 

Geometrically, if the \textit{major} axes\footnote{Major axis of an ellipse is its longer diameter.} of the isolines centered at $\vec{y}$ and $\vec{x}_0$ are aligned in a straight line between $\vec{y}$ and $\vec{x}_0$ as shown in Fig.\;\ref{fig:joint_separate}(b), then: i) $\vec{x}_{MAP}$ \textit{must} reside on the line, and ii) there is only one corresponding observation $\vec{y}$ for a given $\vec{x}_{MAP}$.
The first claim is true because any point not on the straight line between $\vec{y}$ and $\vec{x}_0$ will have a distance sum strictly larger than a point on the straight line and thus cannot be the optimal $\vec{x}_{MAP}$. 
Given the first claim, the second claim is true because $\vec{y}$ must be located at a unique point on an extrapolated line from $\vec{x}_0$ to $\vec{x}_{MAP}$ such that $\vec{x}_{MAP} = \arg \min_{\vec{x}} D_l(\vec{y},\vec{x}) + D_p(\vec{x}_0,\vec{x})$.
In this special case, the solutions to the joint and separate approaches are the same.
Clearly, in the general case (as shown in Fig.\;\ref{fig:joint_separate}(a)) the axes of the isolines are not aligned, and thus in general the joint approach would lead to a better solution. 


\subsection{Rate-Constrained Maximum a Posteriori (MAP) Problem}
\label{subsec:analysis_formulation}

Since the condition of one-to-one correspondence between $\vec{y}$ and $\vec{x}_{MAP}$ is not satisfied in general, we formulate our rate-constrained MAP problem based on the joint approach. 
Instead of (\ref{eq:orig_objective}), one can solve the corresponding Lagrangian relaxed version instead \cite{shoham88}:
\begin{equation}
\label{eq:lagrangian_objective}
\underset{\hat{\vec{x}}\in  \mathcal{S}}{\min}\ -\log P(\hat{\vec{x}}|\vec{y})+\lambda R(\hat{\vec{x}})
\end{equation}
where Lagrangian multiplier $\lambda$ is chosen so that the optimal solution $\hat{\vec{x}}$ to (\ref{eq:lagrangian_objective}) has rate $R(\hat{\vec{x}}) \leq R_{\max}$.
\cite{shoham88,abreu16arxiv} discussed how to select an appropriate $\lambda$.
We focus on how (\ref{eq:lagrangian_objective}) is solved for a given $\lambda > 0$.

We call the first term and the second term in (\ref{eq:lagrangian_objective}) error term and rate term, respectively.
The error term measures, in some sense, the distance between the estimated DCC string $\hat{\vec{x}}$ and the observed DCC string $\vec{y}$.
Note that this error term is not the distance between the estimated DCC string $\hat{\vec{x}}$ and the ground truth DCC string $\vec{x}$, since we have no access to ground truth $\vec{x}$ in practice.
Thus we seek an estimate $\hat{\vec{x}}$ which is somehow close to the observation $\hat{\vec{y}}$, while satisfying some priors of the ground truth signal.
The estimate $\hat{\vec{x}}$ is also required to have small rate term $R(\hat{\vec{x}})$.

Next, we first specify the error term by using a burst error model and a geometric prior in Section \ref{sec:error}.
Then we will specify the rate term using a general variable-length context tree (VCT) model \cite{begleiter2004prediction} in Section \ref{sec:rate}.

\section{Error Term}
\label{sec:error}
\begin{figure}[tb]

\begin{minipage}[b]{.48\linewidth}
  \centering
  \centerline{\includegraphics[width=4cm]{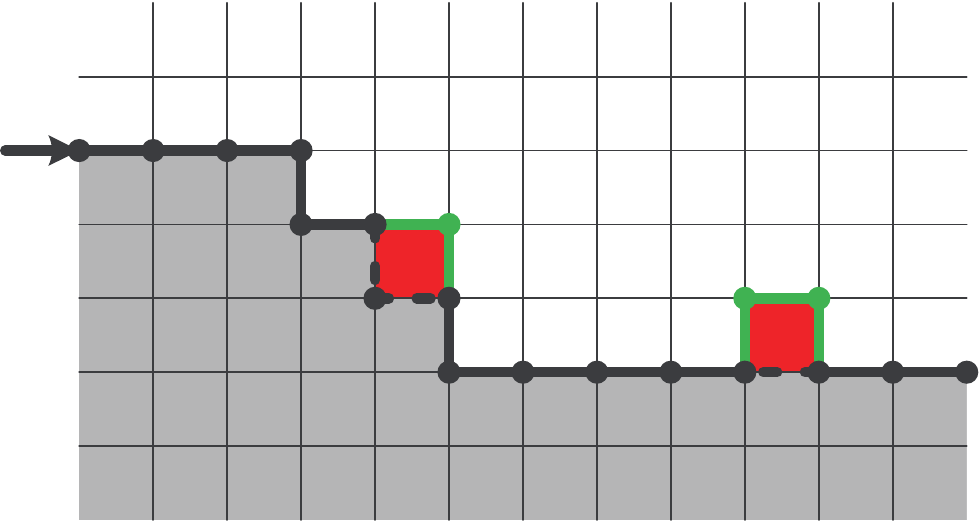}}
  \centerline{(a)}\medskip
\end{minipage}
\hfill
\begin{minipage}[b]{0.48\linewidth}
  \centering
  \centerline{\includegraphics[width=3cm]{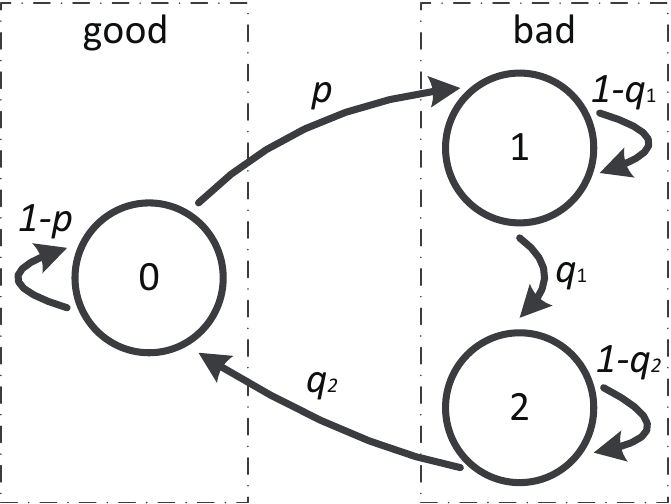}}
  \centerline{(b)}\medskip
\end{minipage}

\vspace{-0.2cm}
\caption{{(a) The two red squares are the erred pixels. Observed $\vec{y}$ is composed of black solid edges (good states) and green solid edges (bad states). The ground truth $\vec{x}$ is composed of black solid edges and black dotted edges. (b) A three-state Markov model.}}
\label{fig:burst_example}
\end{figure}

We first rewrite the posterior $P(\vec{x} | \vec{y})$ in (\ref{eq:lagrangian_objective}) using Bayes' Rule: 
\begin{equation}
P(\vec{x}|\vec{y}) = \frac{P(\vec{y}|\vec{x}) P(\vec{x})}{P(\vec{y})}
\end{equation}
where $P(\vec{y}|\vec{x})$ is the likelihood of observing DCC string $\vec{y}$ given ground truth $\vec{x}$, and $P(\vec{x})$ is the prior which describes \textit{a priori} knowledge about the target DCC string.
We next describe an error model for DCC strings, then define likelihood $P(\vec{y} | \vec{x})$ and prior $P(\vec{x})$ in turn.

%
%
%
%
%
%
%
%

\subsection{Error Model for DCC String}
\label{subsec:error_model}

Assuming that pixels in an image are corrupted by a small amount of independent and identically distributed (iid) noise, a detected contour will occasionally be shifted from the true contour by one or two pixels. 
However, the computed DCC string from the detected contour will experience a sequence of wrong symbols---a \textit{burst error}. 
This is illustrated in Fig.\;\ref{fig:burst_example}(a), where the left single erred pixel (in red) resulted in two erred symbols in the DCC string. The right single error pixel also resulted in a burst error in the observed string, which is \textit{longer} than the original string. Based on these observations, we propose our DCC string error model as follows.

We define a \textit{three-state Markov model} as illustrated in Fig.\;\ref{fig:burst_example}(b) to model the probability of observing DCC string $\vec{y}$ given original string $\vec{x}$. 
State \texttt{0} is the good state, and \textit{burst error state} \texttt{1} and \textit{burst length state} \texttt{2} are the bad states. 
$p$, $q_1$ and $q_2$ are the transition probabilities from state \texttt{0} to \texttt{1}, \texttt{1} to \texttt{2}, and \texttt{2} to \texttt{0}, respectively.
Note that state \texttt{1} cannot transition directly to \texttt{0}, and likewise state \texttt{2} to \texttt{1} and \texttt{0} to \texttt{2}.

Starting at good state \texttt{0}, each journey to state \texttt{1} then to \texttt{2} then back to \texttt{0} is called a \textit{burst error event}.
From state \texttt{0}, each self-loop back to \texttt{0} with probability $1-p$ means that the next observed symbol $y_i$ is the same as $x_i$ in original $\vec{x}$. 
A transition to burst error state \texttt{1} with probability $p$, and each subsequent self-loop with probability $1-q_1$, mean observed $y_i$ is now different from $x_i$.
A transition to burst length state \texttt{2} then models the \textit{length increase} in observed $\vec{y}$ over original $\vec{x}$ due to this burst error event: the number of self-loops taken back to state \texttt{2} is the increase in number of symbols. 
A return to good state \texttt{0} signals the end of this burst error event.

\subsection{Likelihood Term}
\label{subsec:error_likelihod}

Given the three-state Markov model, we can compute likelihood $P(\vec{y} | \vec{x})$ as follows. 
For simplicity, we assume that $\vec{y}$ starts and ends at good state \texttt{0}. 
Denote by $K$ the total number of burst error events in $\vec{y}$ given $\vec{x}$. 
Further, denote by $l_1(k)$ and $l_2(k)$ the number of visits to state \texttt{1} and \texttt{2} respectively during the $k$-th burst error event.
Similarly, denote by $l_0(k)$ the number of visits to state \texttt{0} \textit{after} the $k$-th burst error event. 
We can then write the likelihood $P(\vec{y}|\vec{x})$ as:

\vspace{-0.1in}
\begin{small}
\begin{equation}
\begin{split}
&(1-p)^{l_0(0)} \prod^{K}_{k=1} 
p(1-q_1)^{l_1(k)-1} q_1(1-q_2)^{l_2(k)-1}q_2 (1-p)^{l_0(k)-1} 
\end{split}
\label{eq:likelihood}
\end{equation}
\end{small}

\vspace{-0.15in}
For convenience, we define the total number of visits to state \texttt{0}, \texttt{1} and \texttt{2} as $\Gamma=\sum_{k=0}^{K} l_0(k)$, $\Lambda=\sum_{k=1}^K l_1(k)$ and $\Delta=\sum_{k=1}^K l_2(k)$, respectively.
We can then write the negative log of the likelihood as:
\begin{equation}
\label{eq:negaLogLikelihood}
\begin{split}
&-\log{P(\vec{y}|\vec{x})} =\\
& - K(\log{p}+\log{q_1}+\log{q_2}) -(\Gamma-K)\log{(1-p)} \\
&  - (\Lambda-K)\log{(1-q_1)} - (\Delta-K)\log{(1-q_2)}
\end{split}
\end{equation}

Assuming that burst errors are rare events, $p$ is small and $\log(1-p) \approx 0$. Hence:
\begin{equation}
\label{eq:negaLogLikelihoodAppro}
\begin{split}
& -\log{P(\vec{y}|\vec{x})} \\
& \approx  - K(\log{p}+\log{q_1}+\log{q_2}) \\ 
& - (\Lambda-K)\log{(1-q_1)} - (\Delta-K)\log{(1-q_2)} \\
& = - K(\underbrace{\log{p}+\log{\frac{q_1}{1-q_1}}+\log{\frac{q_2}{1-q_2}}}_{-c_0}) \\
& - \Lambda \underbrace{\log(1-q_1)}_{-c_1} -
\Delta \underbrace{\log(1-q_2)}_{-c_2}
\end{split}
\end{equation}
Thus $-\log P(\vec{y}|\vec{x})$ simplifies to:
\begin{equation}
\label{eq:negaLogLikelihoodApproSimple}
\begin{split}
& -\log{P(\vec{y}|\vec{x})}  \approx  (c_0 + c_2) K + c_1 \Lambda 
+ c_2 \Delta'
\end{split}
\end{equation}
where $\Delta' = \Delta - K$ is the length increase in observed $\vec{y}$ compared to $\vec{x}$ due to the $K$ burst error events\footnote{Length increase of observed $\vec{y}$ due to $k$-th burst error event is $l_2(k) - 1$.}.
(\ref{eq:negaLogLikelihoodApproSimple}) states that the negative log of the likelihood is a linear sum of three terms: i) the number of burst error events $K$; ii) the number of error corrupted symbols $\Lambda$; and iii) the length increase $\Delta'$ in observed string $\vec{y}$.
This agrees with our intuition that more error events, more errors and more deviation in DCC length will result in a larger objective value in (\ref{eq:lagrangian_objective}).
We will validate this error model empirically in our experiments in Section \ref{sec:results}.

\subsection{Prior Term}
\label{subsec:error_prior}

Similarly to \cite{daribo14,zheng17}, we propose a \textit{geometric shape prior} based on the assumption that contours in natural images tend to be more straight than curvy.
Specifically, we write prior $P(\vec{x})$ as:
\begin{equation}
P(\vec{x}) = \exp\left\{- \beta \sum\limits_{i=D_s+1}^{L_{\vec{x}}} s(\vec{x}_{i-D_s}^{i})\right\}
\label{eq:geometric_prior}
\end{equation}
where $\beta$ and $D_s$ are parameters. 
$s(\vec{x}_{i-D_s}^i)$ measures the \textit{straightness} of DCC sub-string $\vec{x}_{i-D_s}^i$.
Let $\vec{w}$ be a DCC string of length $D_s+1$, \textit{i.e.}, $L_{\vec{w}}=D_s+1$. 
Then $s(\vec{w})$ is defined as the \textit{maximum Euclidean distance} between any coordinates of edge $\vec{e}_{\vec{w}}(k), 0\leq k \leq L_{\vec{w}}$ and the line connecting the first point $(m_0,n_0)$ and the last point $(m_{L_{\vec{w}}}, n_{L_{\vec{w}}})$ of $\vec{w}$ on the 2D grid.
We can write $s(\vec{w})$ as:

\vspace{-0.1in}
\begin{small}
\begin{equation}
\label{eq:straight}
\begin{split}
& \underset{0\! \leq k \leq L_{\vec{w}}}{\max}\left\{\!\frac{|(m_k\!-\!m_0)(n_{L_{\vec{w}}}\!-\!n_0)\!-\!(n_k\! -\! n_0)(m_{L_{\vec{w}}}\!-\!m_0) |}{\sqrt{(m_{L_{\vec{w}}}-m_0)^2+(n_{L_{\vec{w}}}-n_0)^2}} \! \right\}
\end{split}
\end{equation}
\end{small}

\begin{figure}[h]

\begin{minipage}[b]{.32\linewidth}
  \centering
  \centerline{\includegraphics[width=1.5cm]{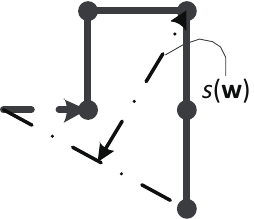}}
  \centerline{(a)}\medskip
\end{minipage}
\hfill
\begin{minipage}[b]{0.32\linewidth}
  \centering
  \centerline{\includegraphics[width=1.3cm]{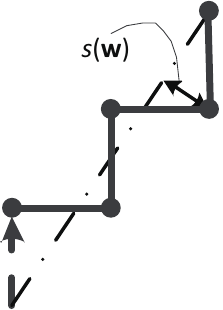}}
  \centerline{(b)}\medskip
\end{minipage}
\hfill
\begin{minipage}[b]{0.32\linewidth}
  \centering
  \centerline{\includegraphics[width=2.8cm]{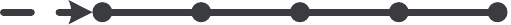}}
  \centerline{(c)}\medskip
\end{minipage}

\vspace{-0.3cm}
\caption{Three examples of the straightness of $s(\vec{w})$ with $L_{\vec{w}}=4$. (a) $\vec{w}=rrls$ and $s(\vec{w})=4\sqrt{5}/5$. (b) $\vec{w}=lrlr$ and $s(\vec{w})=6\sqrt{13}/13$. (c) $\vec{w}=ssss$ and $s(\vec{w})=0$.}
\label{fig:prior}
\end{figure}

Since we compute the sum of a straightness measure for overlapped sub-strings, the length of each sub-string $\vec{x}_{i-D_s}^i$ should not be too large to capture the local contour behavior.
Thus, we choose $D_s$ to be a fixed small number in our implementation. 
Some examples of $s(\vec{w})$ are shown in Fig.\;\ref{fig:prior}.


Combining the likelihood and prior terms, we get the negative log of the posterior,
\begin{equation}
\label{eq:posterior}
\begin{split}
 -\log P(\hat{\vec{x}}|\vec{y})= & (c_0 + c_2) K + c_1 \Lambda + c_2 \Delta' \\
& +\beta \sum\limits_{i=D_s+1}^{L_{\hat{\vec{x}}}} s(\hat{\vec{x}}_{i-D_s}^i)
\end{split}
\end{equation}

\section{Rate Term}
\label{sec:rate}
We losslessly encode a chosen DCC string $\hat{\vec{x}}$ using arithmetic coding \cite{DCC1st1991}. 
Specifically, to implement arithmetic coding using local statistics, each DCC symbol $\hat{x}_i \in \mathcal{A}$ is assigned a conditional probability $P(\hat{x}_i|\hat{\vec{x}}_{1}^{i-1})$ given its all previous symbols $\hat{\vec{x}}_{1}^{i-1}$.
The rate term $R(\hat{\vec{x}})$ is thus approximated as the summation of negative log of conditional probabilities of all symbols in $\hat{\vec{x}}$:
\begin{equation}
\label{eq:rate}
R(\hat{\vec{x}})= - \sum\limits_{i=1}^N \log_2 P(\hat{x}_i|\hat{\vec{x}}_{1}^{i-1}) 
\end{equation}

To compute $R(\hat{\vec{x}})$, one must assign conditional probabilities $P(\hat{x}_i|\hat{\vec{x}}_{1}^{i-1})$ for all symbols $\hat{x}_i$.
In our implementation, we use the variable-length context tree model \cite{begleiter2004prediction} to compute the probabilities.
Specifically, to code contours in the target image, one context tree is trained using contours in a set of training images\footnote{For coding contours in the target frame of a video, the training images are the earlier coded frames.} which have correlated statistics with the target image.
Next we introduce the context tree model, then discuss our construction of a context tree using \textit{prediction by partial matching} (PPM)\footnote{While the computation of conditional probabilities are from PPM, our construction of a context tree is novel and stands as one key contribution in this paper.} \cite{moffat1990implementing}.

\begin{figure}[t]

\begin{minipage}[b]{1\linewidth}
  \centering
  \centerline{\includegraphics[width=8 cm]{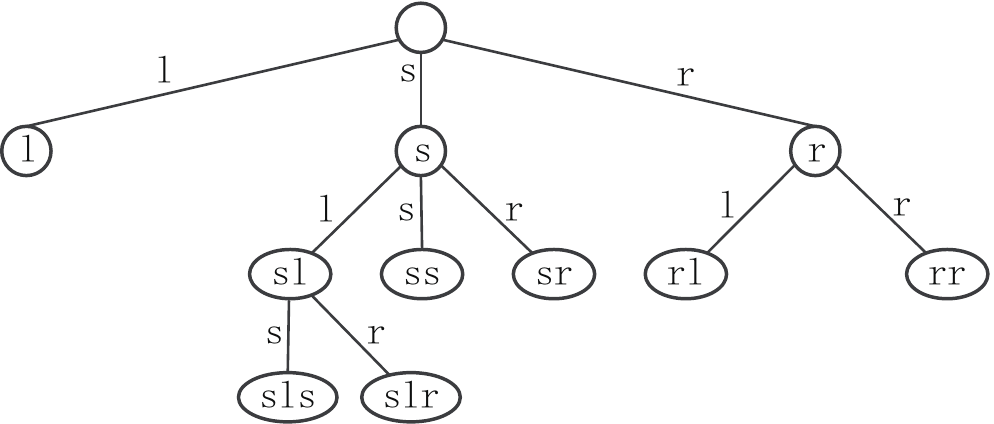}}
  \centerline{}
\end{minipage}

\vspace{-0.2cm}
\caption{An example of context tree.
Each node is a sub-string and the root node is an empty sub-string.
The contexts are all the end nodes on $\mathcal{T}$: $\mathcal{T}=\{\texttt{l},\texttt{sl},\texttt{sls},\texttt{slr},\texttt{ss},\texttt{sr},\texttt{rl},\texttt{r},\texttt{rr}\}$.}
\label{fig:MCT_statespace}
\end{figure}

\subsection{Definition of Context Tree}
\label{subsec:contextTree}

We first define notations related to the contours in the set of training images.
Denote by $\vec{x}(m)$, $1 \leq m \leq M$, the $m$-th DCC string in the training set $\mathcal{X} = \{\vec{x}(1),\ldots,\vec{x}(M)\}$, where $M$ denotes the total number of DCC strings in $\mathcal{X}$. The total number of symbols in $\mathcal{X}$ is denoted by $L = \sum^{M}_{m=1}L_{\vec{x}(m)}$. 
Denote by $\vec{u}\vec{v}$ the concatenation of sub-strings $\vec{u}$ and $\vec{v}$.

We now define $N(\vec{u})$ as the number of occurrences of sub-string $\vec{u}$ in the training set $\mathcal{X}$. $N(\vec{u})$ can be computed as:
\begin{equation}
N(\vec{u})=\sum^{M}_{m=1}\sum_{i=1}^{L_{\vec{x}(m)} - |\vec{u}| + 1}
\vec{1} \left(
\vec{x}(m)_i^{i+|\vec{u}|-1} = \vec{u}
\right) 
\end{equation}
where $\vec{1}(\vec{c})$ is an indicator function that evaluates to $1$ if the specified binary clause $\vec{c}$ is true and $0$ otherwise.

Denote by $P(x|\vec{u})$ the conditional probability of symbol $x$ occurring given its previous sub-string is $\vec{u}$, where $x\in\mathcal{A}$. Given training data $\mathcal{X}$, $P(x|\vec{u})$ can be estimated using $N(\vec{u})$ as done in \cite{buhlmann1999variable},
\begin{equation}
\begin{array}{cc}
P(x|\vec{u})=\frac{N(x\vec{u})}{\delta +  N(\vec{u})}
\end{array}
\label{eq:cal_prob}
\end{equation}
where $\delta$ is a chosen parameter for different models.

Given $\mathcal{X}$, we learn a context model to assign a conditional probability to any symbol given its previous symbols in a DCC string. Specifically, to calculate the conditional probability $P(\hat{x}_i|\hat{\vec{x}}^{i-1}_{1})$, the model determines a \textit{context} $\vec{w}$ to calculate $P(\hat{x}_i|\vec{w})$, where $\vec{w}$ is a \textit{prefix} of the sub-string $\hat{\vec{x}}^{i-1}_{1}$, \textit{i.e.}, $\vec{w} = \hat{\vec{x}}^{i-1}_{i-l}$ for some context length $l$:
\begin{equation}
P(\hat{x}_i|\hat{\vec{x}}^{i-1}_{1}, \text{context model})=P(\hat{x}_i|\vec{w})
\label{eq:conditionalProbContextModel}
\end{equation}
$P(\hat{x}_i|\vec{w})$ is calculated using (\ref{eq:cal_prob}) given $\mathcal{X}$. 
The context model determines a unique context $\vec{w}$ of finite length for every possible past $\hat{\vec{x}}^{i-1}_{1}$. The set of all mappings from $\hat{\vec{x}}^{i-1}_{1}$ to $\vec{w}$ can be represented compactly as a context tree.

Denote by $\mathcal{T}$ the context tree, where $\mathcal{T}$ is a \textit{ternary tree}: each node has at most three children. 
The root node has an empty sub-string, and each child node has a sub-string $\vec{u}x$ that is a concatenation of: i) its parent's sub-string $\vec{u}$ if any, and ii) the symbol $x$ (one of \texttt{l}, \texttt{s} and \texttt{r}) representing the link connecting the parent node and itself in $\mathcal{T}$. 
An example is shown in Fig.\;\ref{fig:MCT_statespace}.
The contexts of the tree $\mathcal{T}$ are the sub-strings of the \textit{context nodes}---nodes that have at most two children, \textit{i.e.}, the end nodes and the intermediate nodes with fewer than three children. 
Note that $\mathcal{T}$ is completely specified by its set of context nodes and vice versa.
For each $\hat{\vec{x}}^{i-1}_{1}$, a context $\vec{w}$ is obtained by traversing $\mathcal{T}$ from the root node to the deepest context node, matching symbols $\hat{x}_{i-1}, \hat{x}_{i-2}, \ldots$ into the past.
We can then rewrite (\ref{eq:conditionalProbContextModel}) as follows:
\begin{equation}
P(\hat{x}_i|\hat{\vec{x}}^{i-1}_{1}, \mathcal{T})=P(\hat{x}_i|\vec{w}).
\label{eq:conditionalProbContextTree}
\end{equation}

\subsection{Construction of Context Tree by Prediction by Partial Matching (PPM)}
\label{subsec:PPM}

The PPM algorithm is considered to be one of the best lossless compression algorithms \cite{begleiter2004prediction}, which is based on the context tree model.
Using PPM, all the possible sub-strings $\vec{w}$ with non-zero occurrences in $\mathcal{X}$, \textit{i.e.}, $N(\vec{w}) > 0$, are contexts on the context tree. 
The key idea of of PPM is to deal with the \textit{zero frequency} problem when estimate $P(\hat{x}_i|\vec{w})$, where sub-string $\hat{x}_i\vec{w}$ does not occur in $\mathcal{X}$, \textit{i.e.}, $N(\hat{x}_i\vec{w})=0$.
In such case, using (\ref{eq:cal_prob}) to estimate $P(\hat{x}_i|\vec{w})$ would result in zero probability, which cannot be used for arithmetic coding.
When $N(\hat{x}_i\vec{w})=0$, $P(\hat{x}_i|\vec{w})$ is estimated instead by reducing the context length by one, \textit{i.e.}, $P(\hat{x}_i|\vec{w}_2^{|\vec{w}|})$.
If sub-string $\hat{x}_i \vec{w}_2^{|\vec{w}|}$ still does not occur in $\mathcal{X}$, the context length is further reduced until symbol $\hat{x}_i$ along with the shortened context occurs in $\mathcal{X}$. 
Let $\mathcal{A}_{\vec{w}}$ be an alphabet in which each symbol along with the context $\vec{w}$ occurs in $\mathcal{X}$, \textit{i.e.}, $\mathcal{A}_{\vec{w}}=\{x|N(x\vec{w}) > 0, x \in \mathcal{A}\}$.
Based on the PPM implemented in \cite{moffat1990implementing}, $P(\hat{x}_i|\vec{w})$ is computed using the following (recursive) equation:
\begin{equation}
P(\hat{x}_i|\vec{w}) = 
\begin{cases}
\frac{N(\hat{x}_i \vec{w})}{|\mathcal{A}_{\vec{w}}| + N(\vec{w})},&\text{if } \hat{x}_i \in \mathcal{A}_{\vec{w}} \\
\frac{|\mathcal{A}_{\vec{w}}|}{|\mathcal{A}_\vec{w}| + N(\vec{w})} \cdot P(\hat{x}_i|\vec{w}_2^{|\vec{w}|}),&\text{otherwise}
\end{cases}.
\label{eq:ppm_prob}
\end{equation}

To construct $\mathcal{T}$, we traverse the training data $\mathcal{X}$ once to collect statistics for all potential contexts.
Each node in $\mathcal{T}$, \textit{i.e.}, sub-string $\mathbf{u}$, has three counters which store the number of occurrences of sub-strings $l\mathbf{u}$, $s\mathbf{u}$ and $r\mathbf{u}$, \textit{i.e.}, $N(l\mathbf{u})$, $N(s\mathbf{u})$ and $N(r\mathbf{u})$.
To reduce memory requirement, we set an upper bound $D$ on the maximum depth of $\mathcal{T}$.
As done in \cite{rissanen1983universal}, we choose the maximum depth of $\mathcal{T}$ as $D=\ceil{\ln{L}/\ln{3}}$, which ensures a large enough $D$ to capture natural statistics of the training data of length $L$.
$\mathcal{T}$ is constructed as described in Algorithm \ref{al:contextTree}.

\begin{algorithm}
\caption{Construction of the Context Tree}
\label{al:contextTree}
\begin{algorithmic}[1]

\State{Initialize $\mathcal{T}$ to an empty tree with only root node}

\For{each symbol $x(m)_i,\vec{x}(m)\in\mathcal{X}, \;i\geq D+1$, from $k=1$ to $k=D$ in order}

\If{there exist a node $\mathbf{u}=\vec{x}(m)_{i-k}^{i-1}$ on $\mathcal{T}$}
	\State{increase the counter $N(x(m)_i\mathbf{u})$ by $1$}
\Else
	\State{add node $\mathbf{u}=\vec{x}(m)_{i-k}^{i-1}$ to $\mathcal{T}$}
\EndIf

\EndFor

\end{algorithmic}
\end{algorithm}

The complexity of the algorithm is $O(D \, L)$.
To estimate the code rate of symbol $\hat{x}_i$, we first find the matched context $\vec{w}$ given past $\hat{\vec{x}}_{i-D}^{i-1}$ by traversing the context tree $\mathcal{T}$ from the root node to the deepest node, \textit{i.e.}, $\vec{w}=\hat{\vec{x}}^{i-1}_{i-|\vec{w}|}$, and then compute the corresponding conditional probability $P(\hat{x}_i|\vec{w})$ using (\ref{eq:ppm_prob}).

In summary, having defined the likelihood, prior and rate terms, our Lagrangian objective (\ref{eq:lagrangian_objective}) can now be rewritten as:
\begin{equation}
\label{eq:objective}
\begin{split}
J(\hat{\vec{x}})=& -\log P(\vec{y}|\hat{\vec{x}})-\beta \log P(\hat{\vec{x}}) + \lambda R(\hat{\vec{x}})\\
\approx &(c_0 + c_2) K + c_1 \Lambda + c_2 \Delta' +\beta \sum\limits_{i=D_s+1}^{L_{\hat{\vec{x}}}} s(\hat{\vec{x}}_{i-D_s}^i)\\
 &- \lambda \sum\limits_{i=D+1}^{L_{\hat{\vec{x}}}} \log_2 P(\hat{x}_i|\hat{\vec{x}}_{i-D}^{i-1}). 
\end{split}
\end{equation}
We describe a dynamic programming algorithm to minimize the objective optimally in Section \ref{sec:algorithm}.


\section{Optimization Algorithm}
\label{sec:algorithm}
\begin{figure*}[!t]


\normalsize
\setcounter{mytempeqncnt}{\value{equation}}
\setcounter{equation}{25}

\begin{small}

\begin{equation}
G_i(\hat{\vec{x}}_{i-D}^{i-1},\vec{e},j-1) = \underset{\hat{x}_i \in \mathcal{A}}{\min}
\begin{cases}
f(\hat{\vec{x}}_{i-D}^{i}) ~+~ 
\mathbf{1}(j < L_{\vec{y}})\; G_{i+1}(\hat{\vec{x}}_{i-D+1}^{i}, v(\vec{e},\hat{x}_i),j) , & \text{if}\;\hat{x}_i = y_j \\
(c_0+c_2) + f(\hat{\vec{x}}_{i-D}^{i}) ~+~ B_{i+1}(\hat{\vec{x}}_{i-D+1}^{i}, v(\vec{e},\hat{x}_i),j), & \text{otherwise}
\end{cases}
\label{eq:good_recursion}
\end{equation}

\begin{equation}
B_i(\hat{\vec{x}}_{i-D}^{i-1},\vec{e},j-1)= \underset{\hat{x}_i \in \mathcal{A}}{\min}
\begin{cases}
c_2(k-j)~+~f(\hat{\vec{x}}_{i-D}^{i}) ~+~ G_{i+1}(\hat{\vec{x}}_{i-D+1}^{i}, v(\vec{e},\hat{x}_i),k), & \text{if}\;\exists k, v(\vec{e},\hat{x}_i) = \vec{e}_{\vec{y}}(k) \\
c_1 ~+~ f(\hat{\vec{x}}_{i-D}^{i}) ~+~ B_{i+1}(\hat{\vec{x}}_{i-D+1}^{i}, v(\vec{e},\hat{x}_i),j), & \text{otherwise}
\end{cases}
\label{eq:bad_recursion}
\end{equation}

\end{small}

\hrulefill

\setcounter{equation}{\value{mytempeqncnt}}

\vspace*{4pt}

\end{figure*}

Before processing the detected contours in the target image, a context tree $\mathcal{T}$ is first computed by PPM using a set of training images as discussed in Section\;\ref{sec:rate}.
The rate term $R(\hat{\vec{x}})$ in (\ref{eq:objective}) is computed using $\mathcal{T}$. 
We describe a dynamic programming (DP) algorithm to minimize (\ref{eq:objective}) optimally, then analyze its complexity. 
Finally, we design a total suffix tree (TST) to reduce the complexity.

\subsection{Dynamic Programming Algorithm}
\label{subsec:algorithm_dynamic}

To simplify the expression in the objective, we define a local cost term $f(\hat{\vec{x}}_{i-D}^{i})$ combining the prior and the rate term as:
\begin{equation}
f(\hat{\vec{x}}_{i-D}^{i}) = \beta s(\hat{\vec{x}}_{i-D_s}^i) - \lambda \log_2 P(\hat{x}_i|\hat{\vec{x}}_{i-D}^{i-1}).
\label{eq:local_cost}
\end{equation}
Note that the maximum depth $D=\ceil{\ln{L}/\ln{3}}$ of $\mathcal{T}$ is much larger than $D_s$.
We assume here that the first $D$ estimated symbols $\hat{\vec{x}}^D_1$ are the observed $\vec{y}^D_1$, and that the last estimated edge is correct, \textit{i.e.}, $\vec{e}_{\hat{\vec{x}}}(L_{\hat{\vec{x}}})=\vec{e}_{\vec{y}}(L_{\vec{y}})$.

\addtocounter{equation}{2}

In summary, our algorithm works as follows. 
As we examine each symbol in observed $\mathbf{y}$, we identify an ``optimal" state traversal through our 3-state Markov model---one that minimizes objective (\ref{eq:objective})---via two recursive functions. The optimal state traversal translates directly to an estimated DCC string $\hat{\vec{x}}$, which is the output of our algorithm.

Denote by $G_i(\hat{\vec{x}}_{i-D}^{i-1},\vec{e},j-1)$ the minimum cost for estimated $\hat{\vec{x}}$ from the $i$-th symbol onwards, given that we are in the good state with a set of $D$ previous symbols (context) $\hat{\vec{x}}^{i}_{i-D}$, and last edge is $\vec{e}$ which is the same as the $(j-1)$-th edge  in observed $\vec{y}$. 
If we select one additional symbol $\hat{x}_i = y_j$, then we remain in the good state, incurring a local cost $f(\hat{\vec{x}}^i_{i-D})$, plus a recursive cost $G_{i+1}(\,)$ for the remaining symbols in string $\hat{\vec{x}}$ due to new context $\hat{\vec{x}}^i_{i-D+1}$. 

If instead we choose one additional symbol $\hat{x}_i \neq y_j$, then we start a new burst error event, incurring a local cost $(c_0 + c_2)$ for the new event, in addition to $f(\hat{\vec{x}}^i_{i-D})$.
Entering bad states, we use $B_{i+1}(\,)$ for recursive cost instead. 

$B_i(\hat{\vec{x}}_{i-D}^{i-1},\vec{e},j-1)$ is similarly computed as $G_i(\hat{\vec{x}}_{i-D}^{i-1},\vec{e},j-1)$, except that if selected symbol $\hat{x}_i$ has no corresponding edge in $\vec{y}$, then we add $c_1$ to account for an additional error symbol. 
If selected symbol $\hat{x}_i$ has a corresponding edge in $\vec{y}$, then this is the end of the burst error event, and we return to good state (recursive call to $G_{i+1}(\,)$ instead). In this case, we must account for the change in length between $\hat{\vec{x}}$ and $\vec{y}$ due to this burst error event, weighted by $c_2$. 
$G_i(\hat{\vec{x}}_{i-D}^{i-1},\vec{e},j-1)$ and $B_i(\hat{\vec{x}}_{i-D}^{i-1},\vec{e},j-1)$ are defined in (\ref{eq:good_recursion}) and (\ref{eq:bad_recursion}), respectively. 
$\mathbf{1}(\mathbf{c})$ is an indicator function as defined earlier. 
$\vec{e}=\{(m,n),d\}$ denotes the $(i-1)$-th edge of $\hat{\vec{x}}$, and $v(\vec{e},\hat{x}_i)$ denotes the next edge given that the next symbol is $\hat{x}_i$. $j-1$ is the index of matched edge in $\vec{y}$.

Because we assume that the denoised DCC string $\hat{\vec{x}}$ is no longer than the observed $\vec{y}$, if $i>L_{\vec{y}}$ or $k-j < 0$, we stop the recursion and return infinity to signal an invalid solution.

\subsection{Complexity Analysis}
\label{subsec:algorithm_complexity}

The complexity of the DP algorithm is bounded by the size of the DP tables times the complexity of computing each table entry. 
Denote by $Q$ the number of possible edge endpoint locations.
Looking at the arguments of the two recursive funcions $G_i(\,)$ and $B_i(\,)$, we see that DP table size is $O(3^D \,  4Q \, L_{\vec{y}}^2)$. 
To compute each entry in $B_i(\,)$, for each $\hat{\vec{x}}_i \in \mathcal{A}$,  one must check for matching edge in $\vec{y}$, hence the complexity is $O(3 L_{\vec{y}})$. 
Hence the total complexity of the algorithm is $O(3^D Q L_{\vec{y}}^3)$, which is polynomial time with respect to $L_{\vec{y}}$.

\begin{figure}[tb]

\begin{minipage}[b]{1\linewidth}
  \centering
  \centerline{\includegraphics[width=8cm]{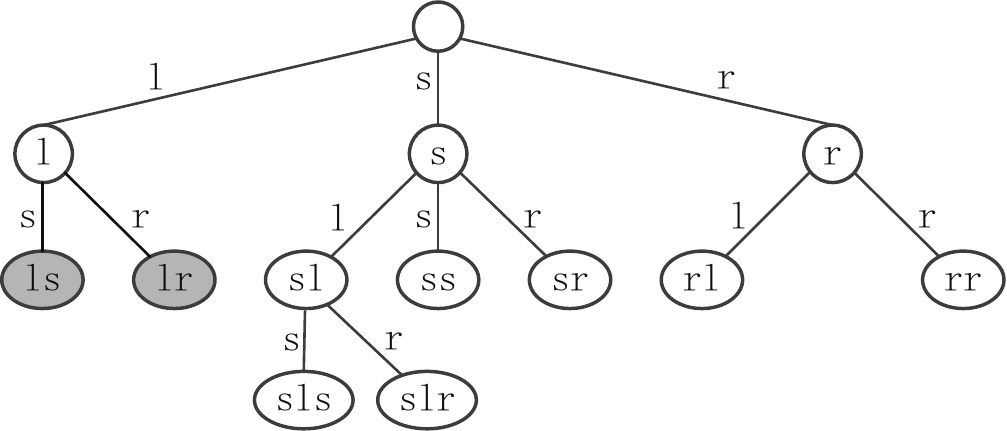}}
  \centerline{}\medskip
\end{minipage}

\vspace{-0.3cm}
\caption{An example of \textit{total suffix tree} (TST) derived from the context tree in Fig.\;\ref{fig:MCT_statespace}. End nodes in gray are added nodes based on the context tree. All the end nodes construct a TST: $\mathcal{T}_s^{\ast}=\{\texttt{l},\texttt{ls},\texttt{lr},\texttt{sl},\texttt{sls},\texttt{slr},\texttt{ss},\texttt{sr},\texttt{rl},\texttt{r},\texttt{rr}\}$.}
\label{fig:suffix_tree}
\end{figure}

\subsection{Total Suffix Tree}
\label{subsec:algorithm_suffix}

When the training data is large, $D$ is also large, resulting in a very large DP table size due to the exponential term $3^D$.
In (\ref{eq:good_recursion}) and (\ref{eq:bad_recursion}) when calculating local cost $f(\hat{\vec{x}}_{i-D}^{i})$, actually the context required to compute rate is $\mathbf{w}=\hat{\mathbf{x}}_{i-|\mathbf{w}|}^{i-1}$, where the context length $|\mathbf{w}|$ is typically smaller than $D$ because the context tree $\mathcal{T}$ of maximum depth $D$ is variable-length.
Thus, if we can, at appropriate recursive calls, reduce the ``history" from $\hat{\mathbf{x}}^{i}_{i+1-D}$ of length $D$ to $\hat{\mathbf{x}}^{i}_{i+1-k}$ of length $k$, $k < D$, for recursive call to $G_{i+1}()$ or $B_{i+1}$ in (\ref{eq:good_recursion}) or (\ref{eq:bad_recursion}), then we can reduce the DP table size and in turn the computation complexity of the DP algorithm.

The challenge is how to retain or ``remember" just enough previous symbols $\hat{x}_{i}, \hat{x}_{i-1}, \ldots$ during recursion so that the right context $\mathbf{w}$ can still be correctly identified to compute rate at a later recursive call.
The solution to this problem can be described simply.
Let $\mathbf{w}$ be a context (context node) in context tree $\mathcal{T}$.
A \textit{suffix} of length $k$ is the first $k$ symbols of $\mathbf{w}$, \textit{i.e.}, $\mathbf{w}^k_1$.
Context $\mathbf{w}$ at a recursive call must be a concatenation of a chosen $i$-th symbol $\hat{x}_i = w_{|\mathbf{w}|}$ during a previous recursive call $G_i()$ or $B_i()$ and suffix $\mathbf{w}^{|\mathbf{w}|-1}_1$.
It implies that suffix $\mathbf{w}^{|\mathbf{w}|-1}_1$ must be retained during the recursion in (\ref{eq:good_recursion}) or (\ref{eq:bad_recursion}) for this concatenation to $\mathbf{w}$ to take place at a later recursion.
To concatenate to suffix $\mathbf{w}^{|\mathbf{w}|-1}_1$ at a later recursive call, one must retain its suffix $\mathbf{w}^{|\mathbf{w}|-2}_1$ at an earlier call.
We can thus generalize this observation and state that \textit{a necessary and sufficient condition to preserve all contexts $\mathbf{w}$ in context tree $\mathcal{T}$ is to retain all suffixes of $\mathbf{w}$ during the recursion.}

All suffixes of contexts in $\mathcal{T}$ can be drawn collectively as a tree as follows. 
For each suffix $\mathbf{s}$, we trace $\mathbf{s}$ down from the root of a tree according to the symbols in $\mathbf{s}$, creating additional nodes if necessary. 
We call the resulting tree a \textit{total suffix tree} (TST)\footnote{An earlier version of TST was proposed in \cite{zheng17} for a different contour coding application.}, denoted as $\mathcal{T}_s$.
By definition, $\mathcal{T}$ is a sub-tree of $\mathcal{T}_s$.
Further, $\mathcal{T}_s$ is essentially a concatenation of all sub-trees of $\mathcal{T}$ at root. 
Assuming $\mathcal{T}$ has $K$ contexts, each of maximum length $D$. 
Each context can induce $O(D)$ additional context nodes in TST $\mathcal{T}_s$. 
Hence TST $\mathcal{T}_s$ has $O(K D)$ context nodes.

Fig.\;\ref{fig:suffix_tree} illustrates one example of TST derived from the context tree shown in Fig.\;\ref{fig:MCT_statespace}.
TST $\mathcal{T}_s$ can be used for compact DP table entry indexing during recursion (\ref{eq:good_recursion}) and (\ref{eq:bad_recursion}) as follows. 
When an updated history $\hat{\mathbf{x}}_{i+1-D}^{i}$ is created from a selection of symbol $\hat{x}_i$, we first truncate $\hat{\mathbf{x}}_{i+1-D}^{i}$ to $\hat{\mathbf{x}}_{i+1-k}^{i}$, where $\hat{\mathbf{x}}_{i+1-k}^{i}$ is the longest matching string in $\mathcal{T}_s$ from root node down. 
The shortened history $\hat{\mathbf{x}}_{i+1-k}^{i}$ is then used as the new argument for the recursive call.
Practically, it means that only DP table entries of arguments $\hat{\mathbf{x}}_{i+1-k}^i$ that are context nodes of TST $\mathcal{T}_s$ will be indexed.
Considering the complexity induced by the prior term, the complexity is thus reduced from original $O(3^D Q L_{\vec{y}}^3)$ to $O( (K D+ 3^{D_s}) Q L_{\vec{y}}^3)$, which is now polynomial in $D$.

\section{Experimental Results}
\label{sec:results}
\subsection{Experimental Setup}
\label{subsec:experimentalSetup}

\begin{figure}[t]

\begin{minipage}[b]{.48\linewidth}
  \centering
  \centerline{\includegraphics[width=4.2cm]{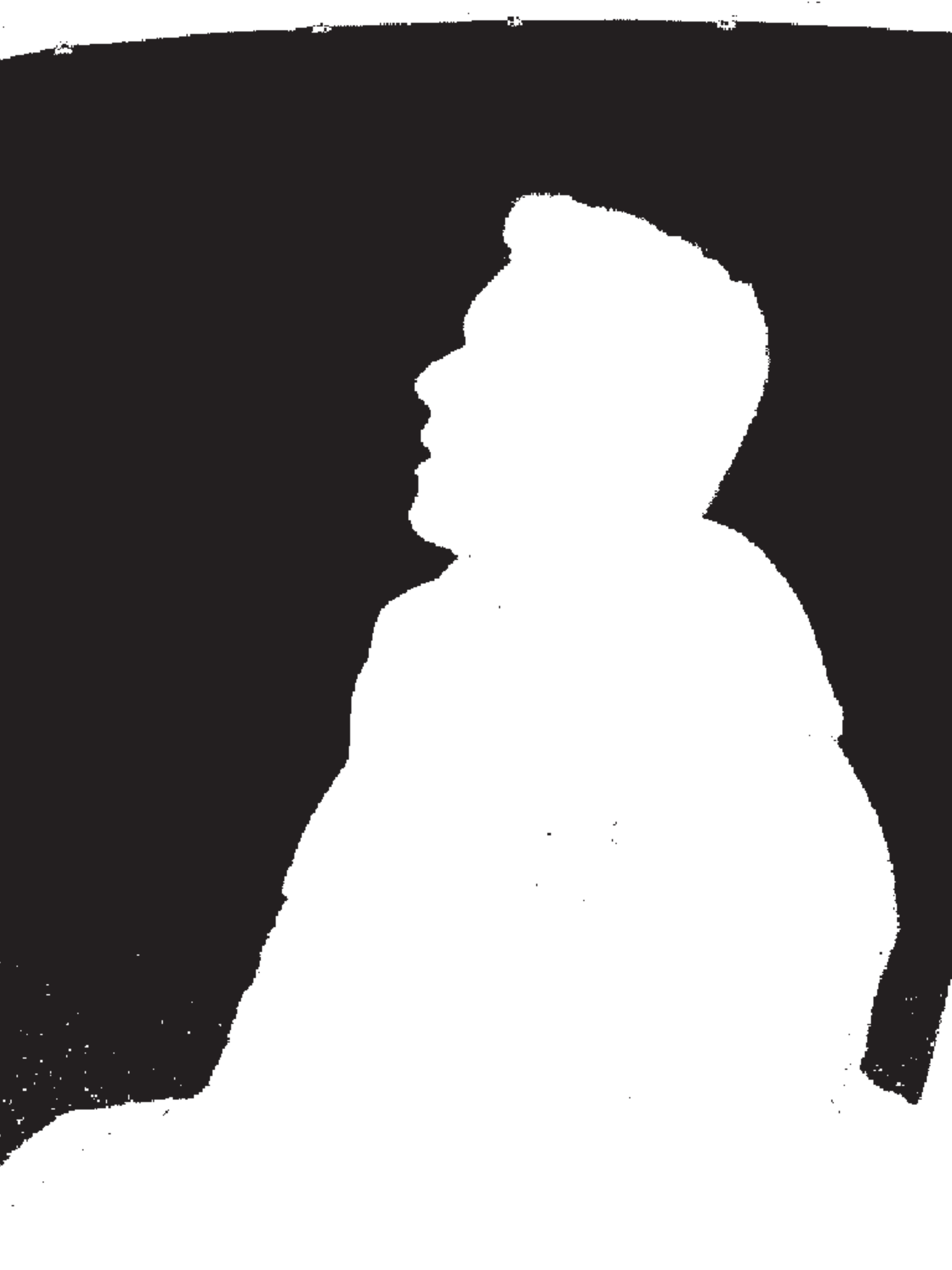}}
  \centerline{(a) \texttt{Model1}}\medskip
\end{minipage}
\hfill
\begin{minipage}[b]{0.48\linewidth}
  \centering
  \centerline{\includegraphics[width=4.2cm]{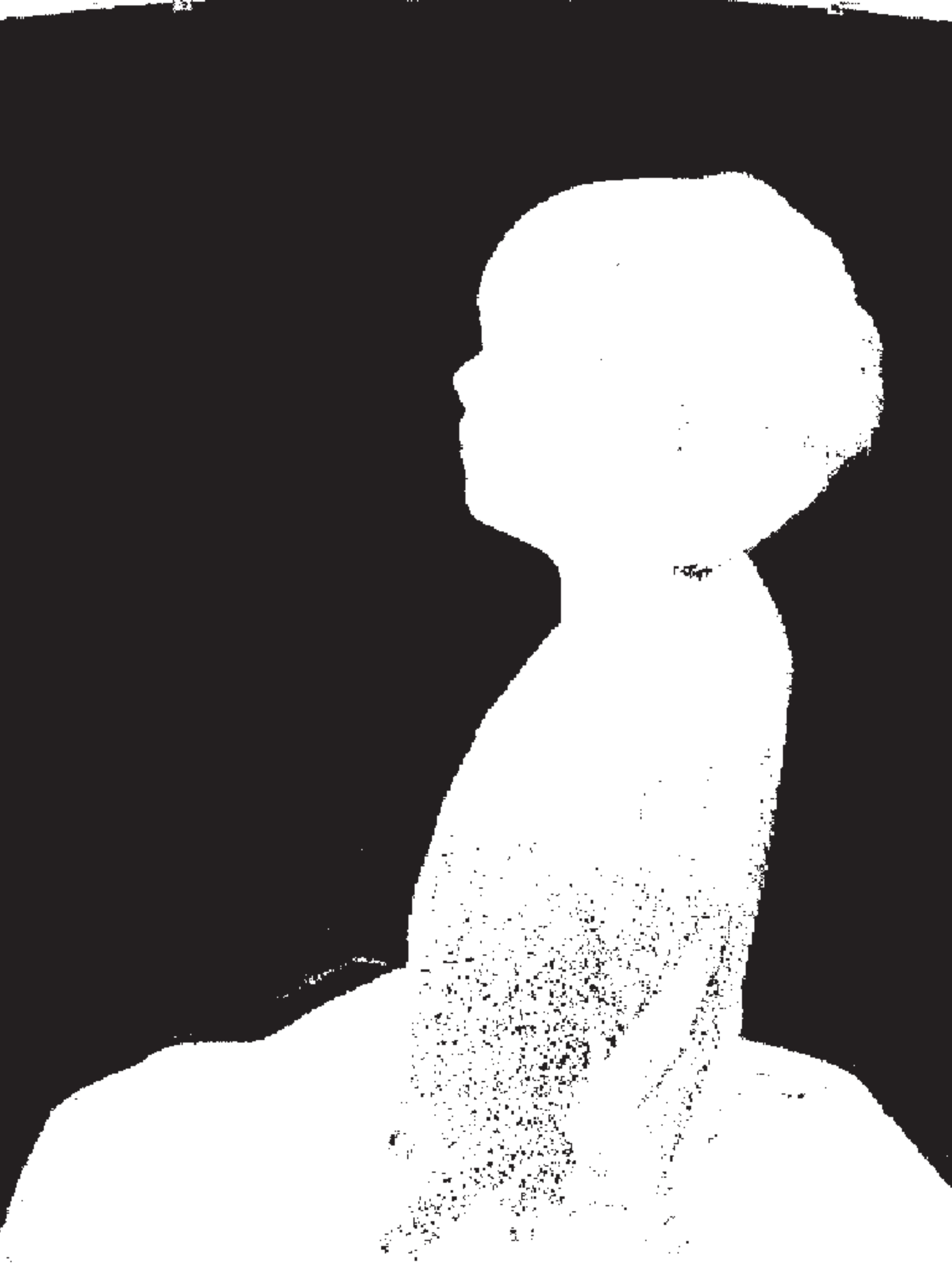}}
  \centerline{(b) \texttt{Model2}}\medskip
\end{minipage}

\vspace{-0.2cm}
\caption{One view of the extracted multiview silhouette sequences.}
\label{fig:silhouette}
\end{figure}

We used two computer-generated (noiseless) depth sequences: \texttt{Dude} (800$\times$400) and \texttt{Tsukuba} (640$\times$480), and two natural (noisy) multiview silhouette sequences: \texttt{Model1} (768$\times$1024) and \texttt{Model2} (768$\times$1024) from Microsoft Research.
10 frames were tested for each sequence.
The multiview silhouette sequences are extracted using the equipment setup in \cite{loop2013real}, where eight views are taken for one frame of each sequence.
The extracted silhouettes are compressed and transmitted for further silhouette-based 3D model reconstruction \cite{mulayim2003silhouette,loop2013real}.
Fig.\;\ref{fig:silhouette} shows one view of \texttt{Model1} and \texttt{Model2}.

We used gradient-based edge detection \cite{daribo14} to detect contours.
For the silhouette sequences, the detected contours were noisy at acquisition.
Note that before detecting the contours for silhouette sequences, we first used a median filter\footnote{http://people.clarkson.edu/~hudsonb/courses/cs611/} to remove the noise inside and outside the silhouette.
For the computer-generated depth images, we first injected noise to the depth images assuming that the pixels along the contours were corrupted by iid noise: for each edge of the contour, the corruption probability was fixed at $\delta$.
If corrupted, the pixel along one side of this edge was replaced by the pixel from the other side (side was chosen with equal probability).
The noisy contours were then detected from the noisy depth images.
Note that for the computer-generated depth images, the ground truth contours were also detected from the original depth images.
We tested two different noise probabilities, $\delta=10\%$ and $\delta=30\%$.

To code contours in a given frame of a depth / silhouette sequence, previous two frames of the same sequence were used to train the context tree.
Unless specified otherwise, we set $D_s=4$ and $\beta = 3$ in (\ref{eq:geometric_prior}) for all experiments.

The quality of each decoded contour was evaluated using \textit{sum of squared distance distortion metric} as proposed in \cite{katsaggelos1998mpeg}: the sum of squared minimum distance from the points on each decoded contour $\vec{\hat{x}}$ to the ground truth contour $\vec{x}$, denoted as $D(\hat{\vec{x}}, \vec{x})$,
\begin{equation}
D(\hat{\vec{x}},\vec{x}) = \sum_{i=1}^{L_{\hat{\vec{x}}}} d^2(\vec{e}_{\hat{\vec{x}}}(i), \vec{x}).
\end{equation}
$d(\vec{e}_{\hat{\vec{x}}}(i), \vec{x})$ is the minimum absolute distance between coordinate $(\hat{m}, \hat{n})$ of the $i$-th edge $\vec{e}_{\hat{\vec{x}}}(i)$ of $\hat{\vec{x}}$ and the segment derived by all edges of $\vec{x}$,
\begin{equation}
d(\vec{e}_{\hat{\vec{x}}}(i), \vec{x}) = \underset{1\leq j \leq L_{\vec{x}}}{\min} |\hat{m}_i-m_j|+|\hat{n}_i-n_j|
\end{equation}
where $(m_j,n_j)$ is the coordinate of the $j$-th edge $\vec{e}_{\vec{x}}(j)$ of $\vec{x}$. 
The main motivation for considering this distortion metric stems from the popularity of the sum of squared distortion measure, which measures the aggregate errors.
Note that some other distortion metrics can also be adopted according to different applications, such as the absolute area between the decoded and ground truth contours, or the depth values change due to the decoded contour shifting compared to the ground truth contour.
For the silhouette sequences without ground truth contours, we set $\lambda=0$ in (\ref{eq:objective}) to get the best denoising result and use it as the ground truth.

We compared performance of four different schemes.
The first scheme, \texttt{Gaussian-ORD}, first denoised the contours using a Gaussian filter \cite{zhong2010convergence}, then encoded the denoised contours using a vertex based lossy contour coding method \cite{lai2010arbitrary}.
The second scheme, \texttt{Lossy-AEC}, first denoised the noisy contours using an irregularity-detection method \cite{daribo14}, then encoded the denoised contours using a DCC based lossy contour coding method \cite{yuan2015contour}.
The third scheme, \texttt{Separate}, first denoised the contours using our proposal by setting $\lambda=0$, then used PPM to encoded the denoised contours.
The fourth scheme, \texttt{Joint}, is our proposal that performed joint denoising / compression of contours.

\subsection{Transition Probability Estimation}
\label{subsec:paraneterEstimation}

\begin{figure}[t]

\begin{minipage}[b]{.48\linewidth}
  \centering
  \centerline{\includegraphics[width=4.5cm]{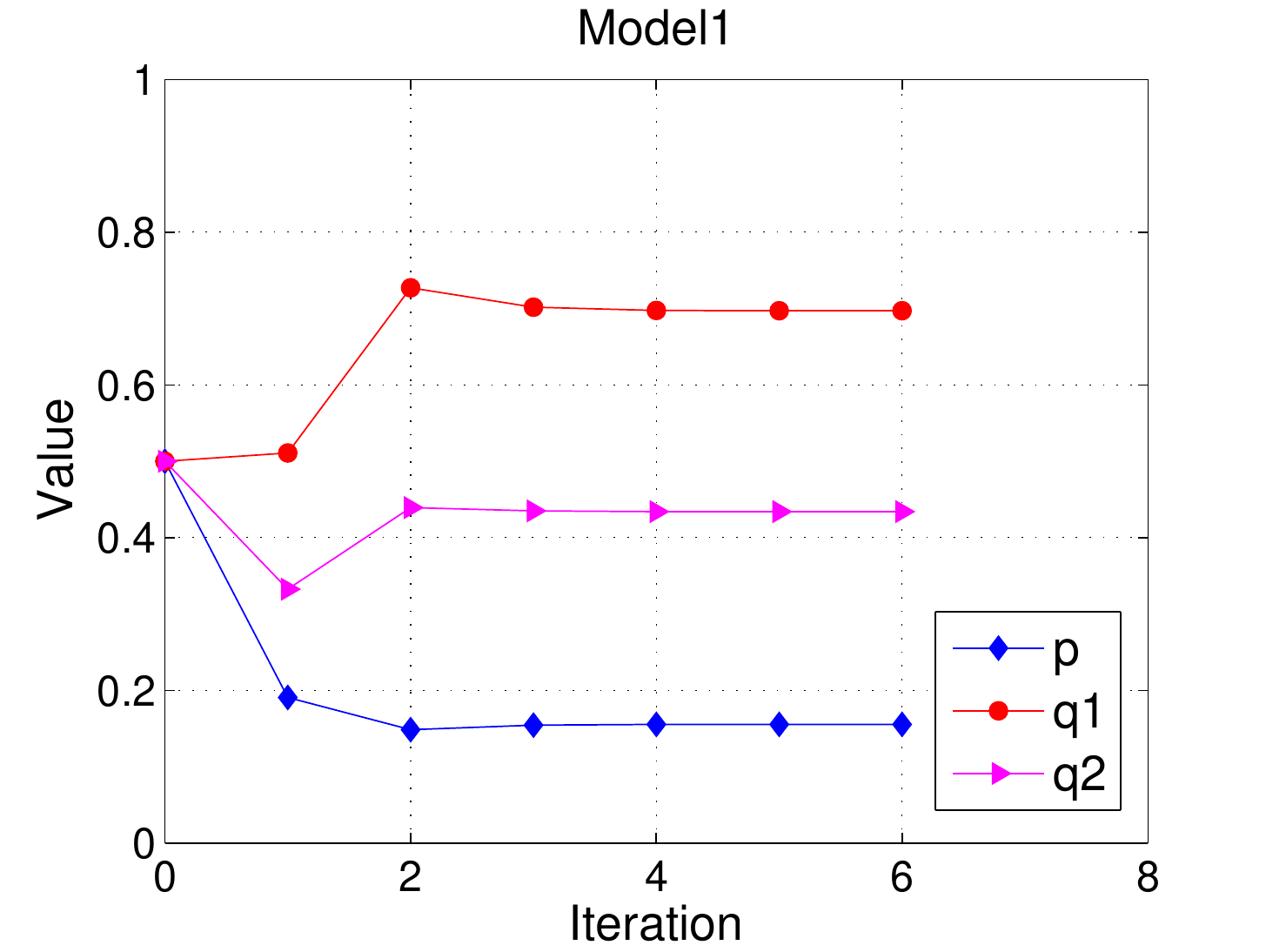}}
\end{minipage}
\hfill
\begin{minipage}[b]{0.48\linewidth}
  \centering
  \centerline{\includegraphics[width=4.5cm]{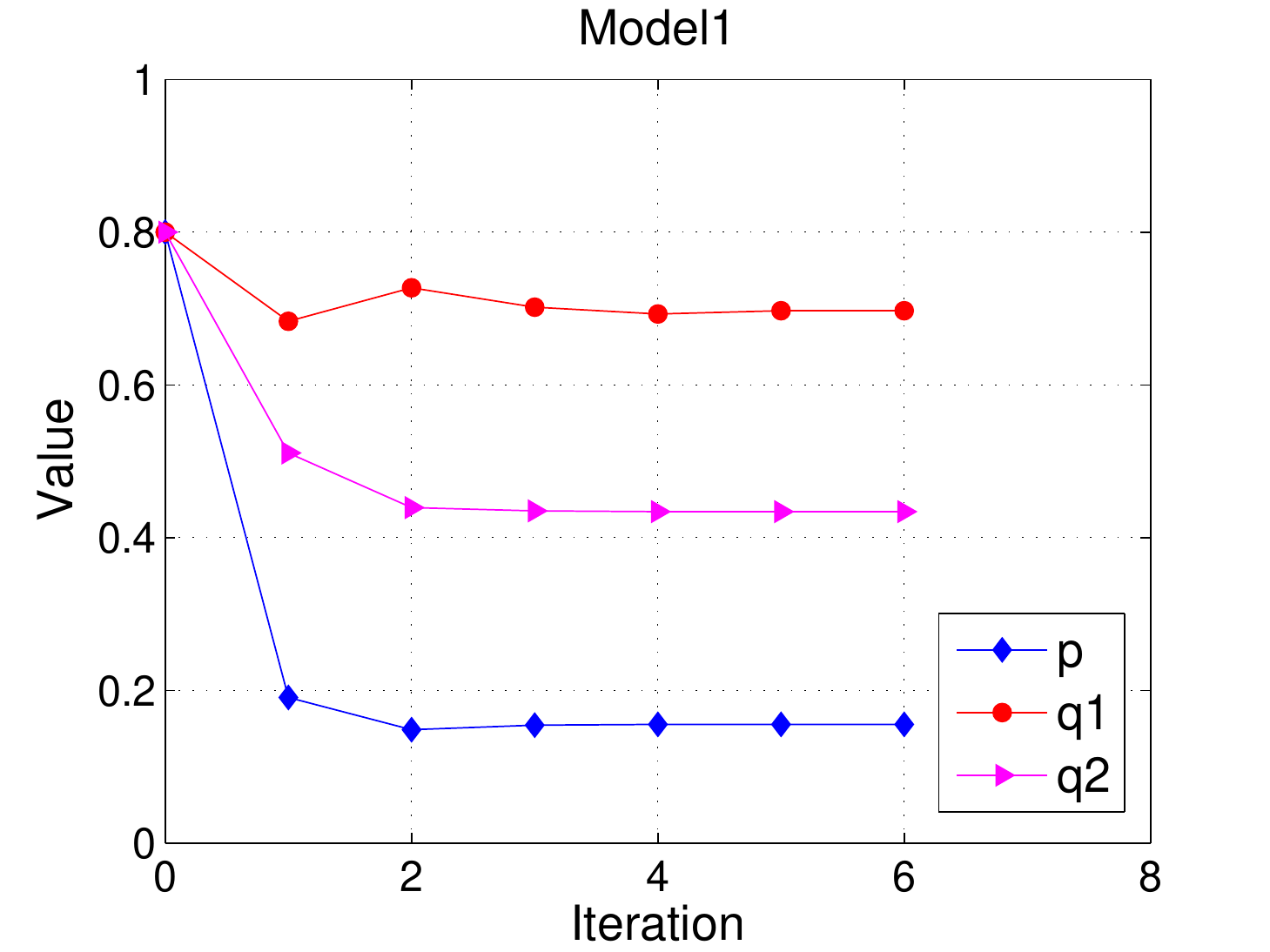}}
\end{minipage}
\vfill
\vspace{0.3cm}
\begin{minipage}[b]{.48\linewidth}
  \centering
  \centerline{\includegraphics[width=4.5cm]{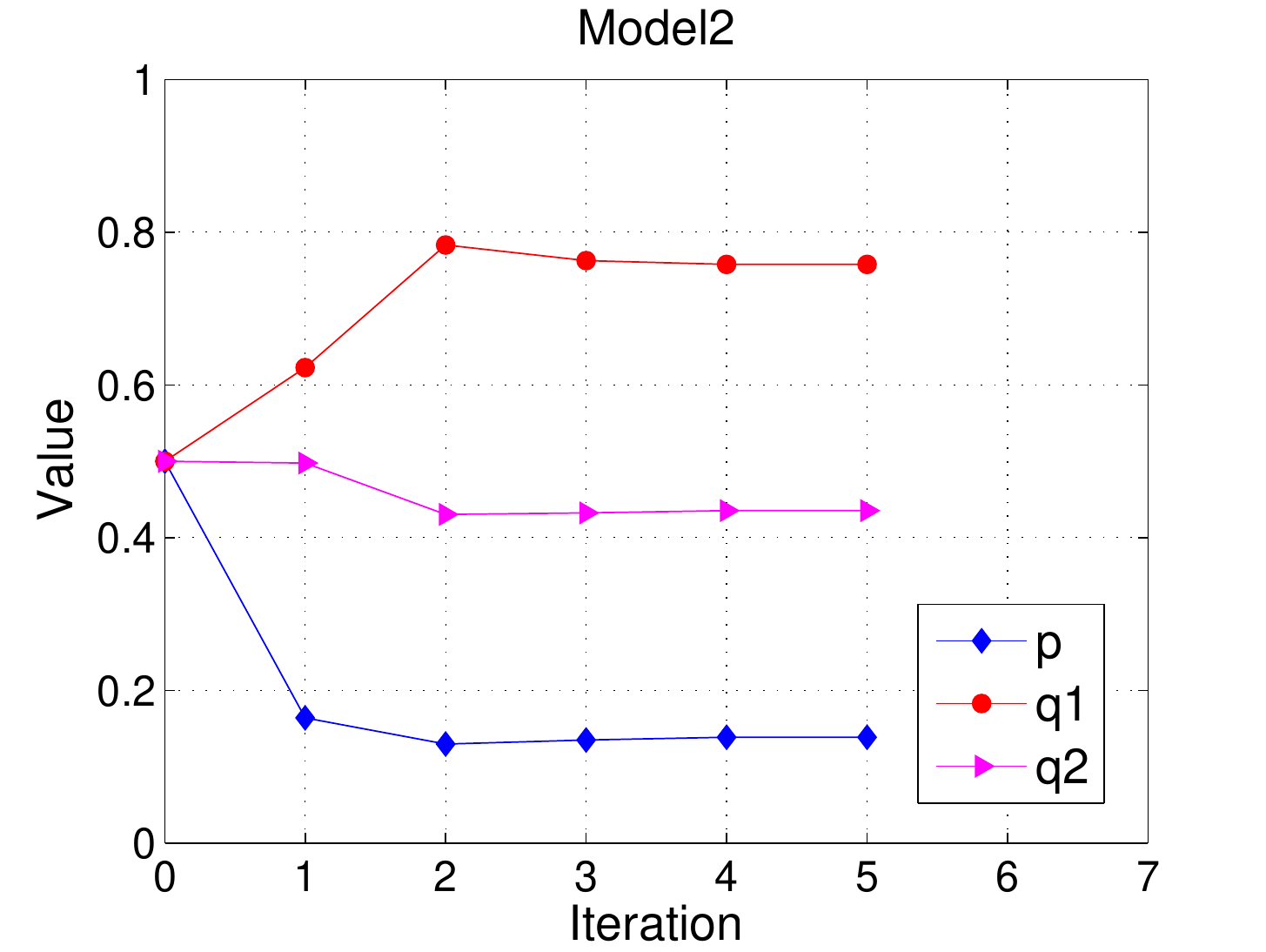}}
\end{minipage}
\hfill
\begin{minipage}[b]{0.48\linewidth}
  \centering
  \centerline{\includegraphics[width=4.5cm]{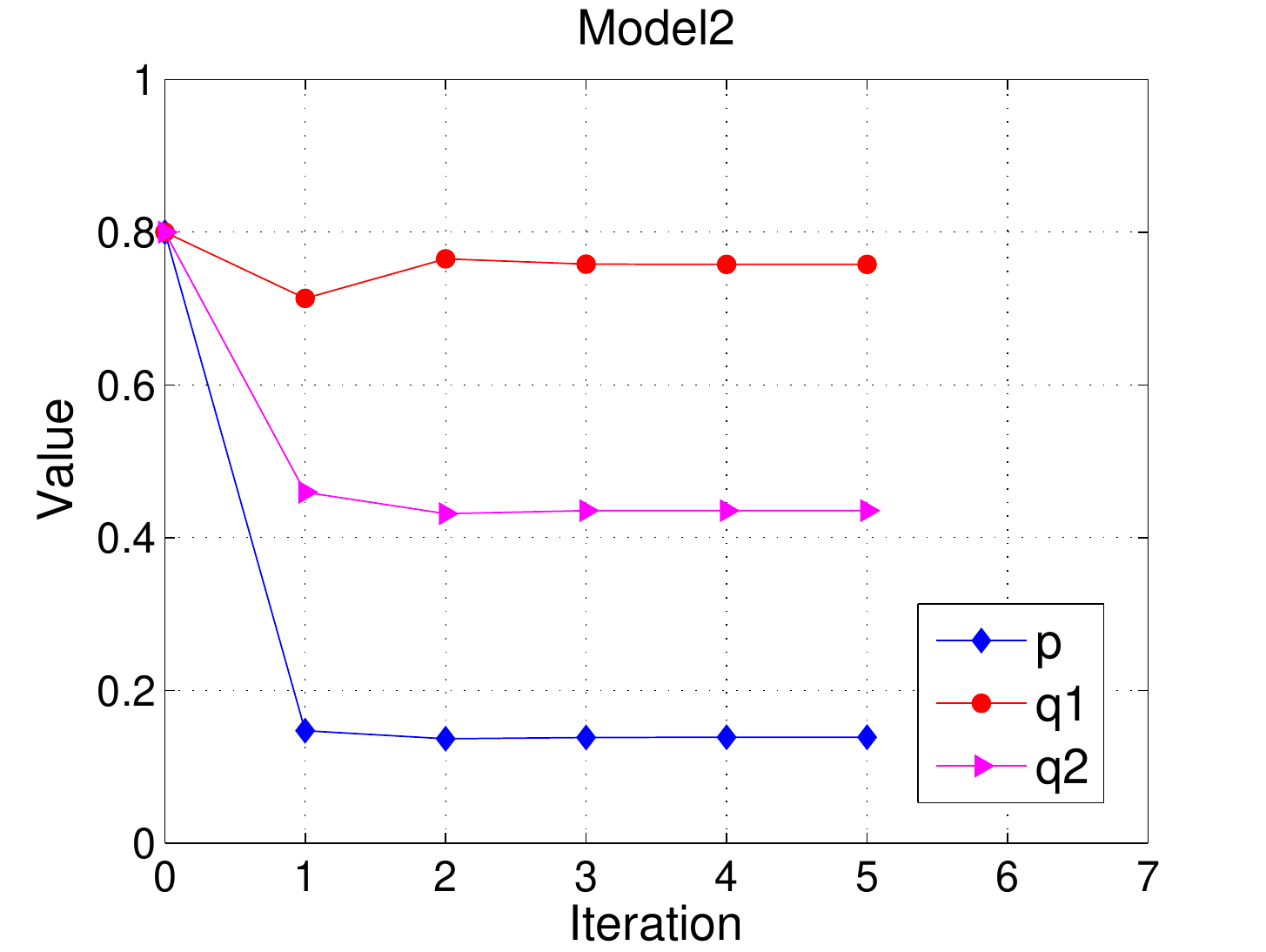}}
\end{minipage}

\caption{Value changes of $p$, $q1$ and $q2$ over number of iterations. Left column: all initial values are setted to 0.5. Right column: all initial values are setted to 0.8.}
\label{fig:parameter_estiamtion}
\end{figure}

For each computer-generated depth sequence, the three transition probabilities in our three-state Markov model, \textit{i.e.}, $p$, $q_1$ and $q_2$, were estimated from average noise statistics by comparing the noisy contours and the ground truth contours.
As described in Section \ref{subsec:error_model}, we traversed the noisy DCC string $\vec{y}$ along with the corresponding ground truth DCC string $\vec{x}$ to count the number of transitions between the three states, \textit{i.e.}, \texttt{0}, \texttt{1} and \texttt{2}.
The transition probability from state\footnote{$j$ can be the same as $i$.} $i$ to state $j$ is thus computed as the number of transitions from state $i$ to state $j$ divided by the number of transitions started from state $i$.

For the silhouette sequences without ground truth contours, we estimated the transition probabilities via an alternating procedure, commonly used when deriving model parameters without ground truth data in machine learning \cite{mackay2003example,sundberg1976iterative}. 
Specifically, given a noisy DCC string $\vec{y}$, we first assigned initial values to the transition probabilities and used them to denoise $\vec{y}$ to get a MAP solution $\vec{x}_{MAP}^{\prime}$ ($\lambda=0$ in (\ref{eq:objective})).
Treating $\vec{x}_{MAP}^{\prime}$ as ground truth and comparing $\vec{y}$ and $\vec{x}_{MAP}^{\prime}$, we updated the transition probabilities.
Then, we used the updated transition probabilities to denoise $\vec{y}$ again to get a new MAP solution and computed the new transition probabilities by comparing $\vec{y}$ and the new MAP solution.
This process was repeated until the transition probabilities converged.

Fig.\;\ref{fig:parameter_estiamtion} shows the changes in model parameters $p$, $q_1$ and $q_2$ across the iterations of the alternating procedure using different initial values for \texttt{Model1} and \texttt{Model2}.
We see that for each silhouette sequence with different initial values, the parameters converged to the same values in only a few iterations.
Note that we can also use the alternating procedure to estimate model parameters for the computer-generated depth sequences, but the estimated results would not be as accurate as those  directly computed using the available ground truth contours and the noisy contours.

\subsection{Proposed Three-state Markov Model versus iid Model}
\label{subsec:vs_iid_model}

To validate our proposed three-state Markov model, we compared to a na\"ive iid generative model for computing the likelihood.
The model is as follows: for each symbol $x_i$ in the ground truth contour $\vec{x}$, we use an iid model (a coin toss) to see if it is erred in the observed contour $\vec{y}$.
If so, we use a Poisson distribution to model the length increase in $\vec{y}$ over $\vec{x}$.
Then we examine the next symbol $x_{i+1}$ and so on until the end of the DCC string.
This model assumes each symbol $x_i$ is independent and is capable of modelling the length increase.

We adopt the widely used Akaike information criterion (AIC) \cite{akaike1974new} to compare the two models.
AIC is a measure of the relative quality of statistical models for a given set of data,
\begin{equation}
AIC = 2k - 2\ln {\mathcal{L}}
\label{eq:AIC}
\end{equation}
where $\mathcal{L}=P(\vec{y}|\vec{x})$ is the likelihood and $k$ is the number of parameters to be estimated.
For the proposed three-state Markov model with three transition probabilities ($p$, $q_1$, $q_2$), $k = 3$; for the i.i.d model, $k=2$ which contains $p^{\prime}$---the probability of each symbol $x_i$ being an error and $\lambda^{\prime}$---the average increased length per erred symbol $x_i$ in Poisson distribution.
$p^{\prime}$ and $\lambda^{\prime}$ were estimated from average noise statistics as similar in Section \ref{subsec:paraneterEstimation}.
To accurately compute AIC, we only test sequence \texttt{Dude} and \texttt{Tsukuba} whose ground truth contours are available.

\begin{table}
\caption{Results of Model Validation}
\label{tab:model_validation}

\begin{tabular}{c|c|c|c|c|c|c}
\hline 
\hline
\multirow{2}{*}{Sequence} & \multirow{2}{*}{Noise} & \multicolumn{2}{c|}{-Log-likelihood} & \multicolumn{2}{c|}{AIC} & \multicolumn{1}{c}{Relative }\tabularnewline
\cline{3-6} 
 &  & i.i.d & Prop & i.i.d & Prop & likelihood\tabularnewline
\hline 
\multirow{2}{*}{Tsukuba} & 10\% & 67 & \textbf{64 } & 134  & \textbf{128 } & 3.93e-02 \tabularnewline
\cline{2-7} 
 & 30\% & 141  & \textbf{134 } & 283  & \textbf{267 } & 3.72e-04 \tabularnewline
\hline 
\multirow{2}{*}{Dude} & 10\% & 198  & \textbf{178 } & 397  & \textbf{356 } & 1.65e-09 \tabularnewline
\cline{2-7} 
 & 30\% & 378 & \textbf{348 } & 756  & \textbf{696 } & 8.93e-14 \tabularnewline
\hline 
\hline
\end{tabular}

\end{table}

Table\;\ref{tab:model_validation} shows the results of the negative log-likelihood and AIC of the two models.
All the values of negative log-likelihood and AIC are averaged to one contour for better comparison. 
A smaller value of AIC represents a better fit of the model.
We can see that the proposed three-state Markov model always achieved smaller values.
Relative likelihood \cite{burnham2003model}, computed as $exp((AIC_{prop}-AIC_{iid})/2)$, in the last column is interpreted as being proportional to the probability that the iid model is the proposed model.
For example, for $Tsukuba$ with 10\% noise, it means the iid model is 3.93e-2 times as probable as the proposed model to minimize the information loss.
All the results show that the proposed three-state Markov model fits the data better than the iid model.

\subsection{Performance of Rate Distortion}
\label{subsec:RD_performance}

\begin{figure}[t]

\begin{minipage}[b]{.48\linewidth}
  \centering
  \centerline{\includegraphics[width=4.5cm]{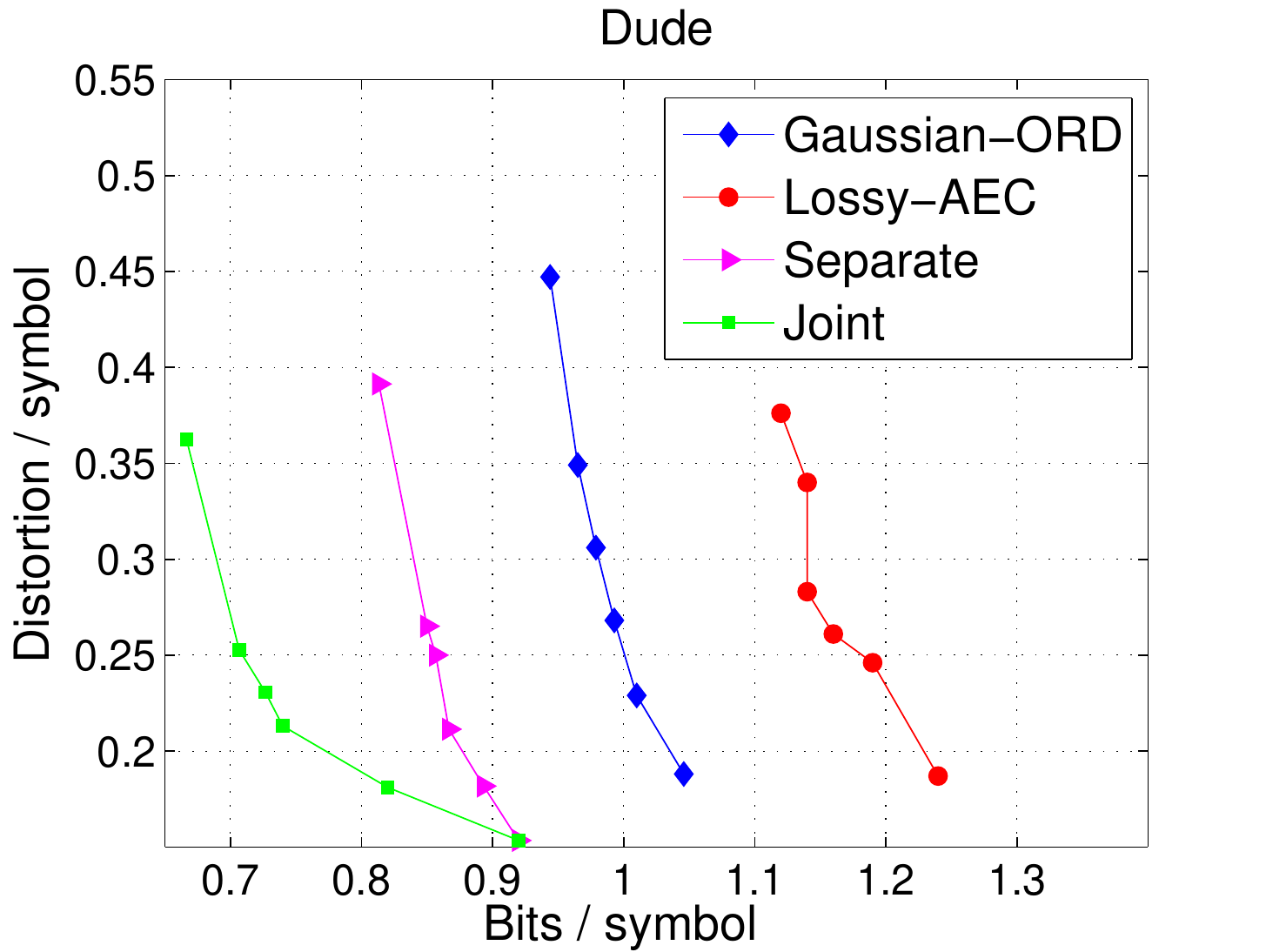}}
\end{minipage}
\hfill
\begin{minipage}[b]{0.48\linewidth}
  \centering
  \centerline{\includegraphics[width=4.5cm]{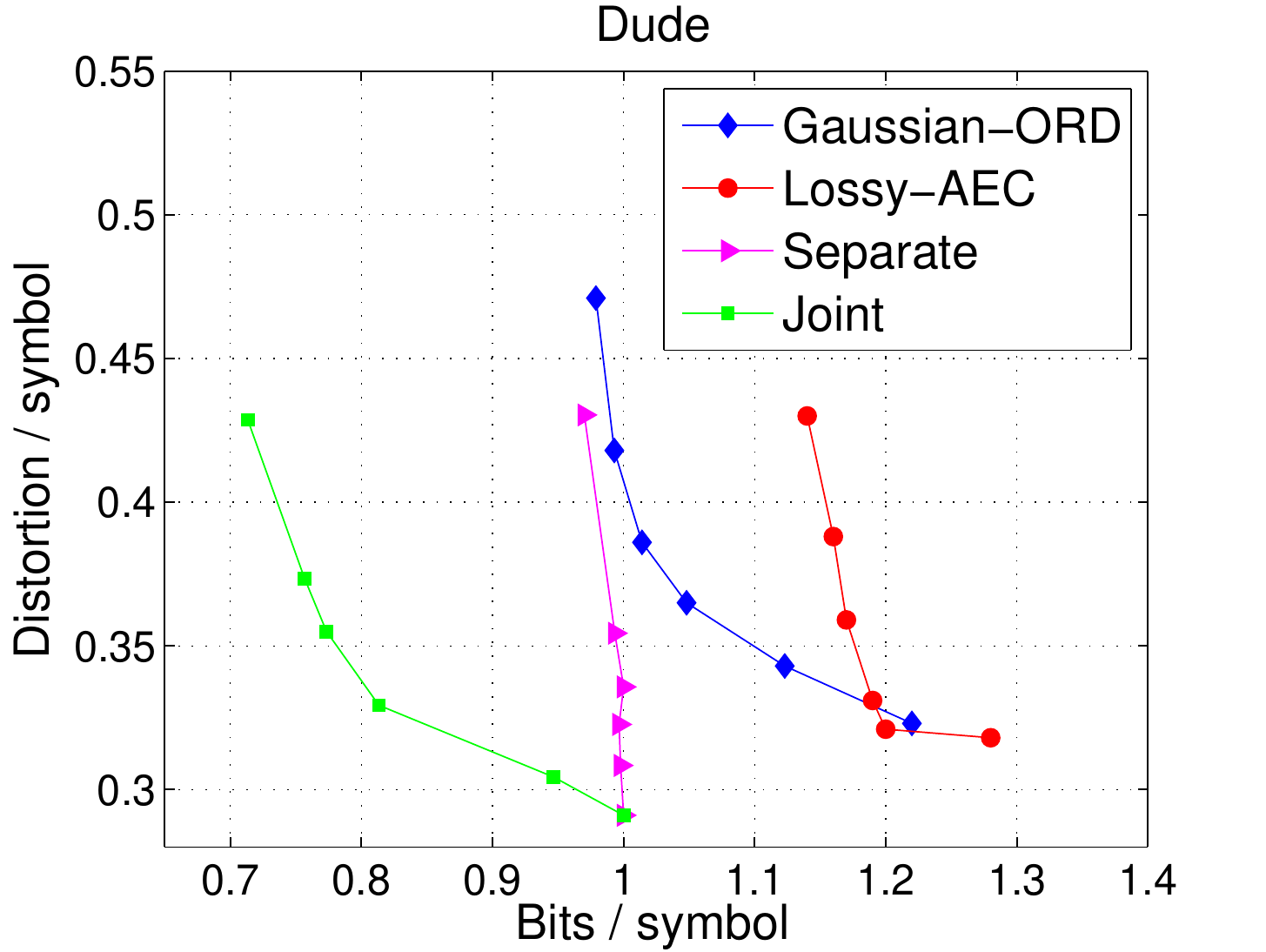}}
\end{minipage}
\vfill
\vspace{0.3cm}
\begin{minipage}[b]{.48\linewidth}
  \centering
  \centerline{\includegraphics[width=4.5cm]{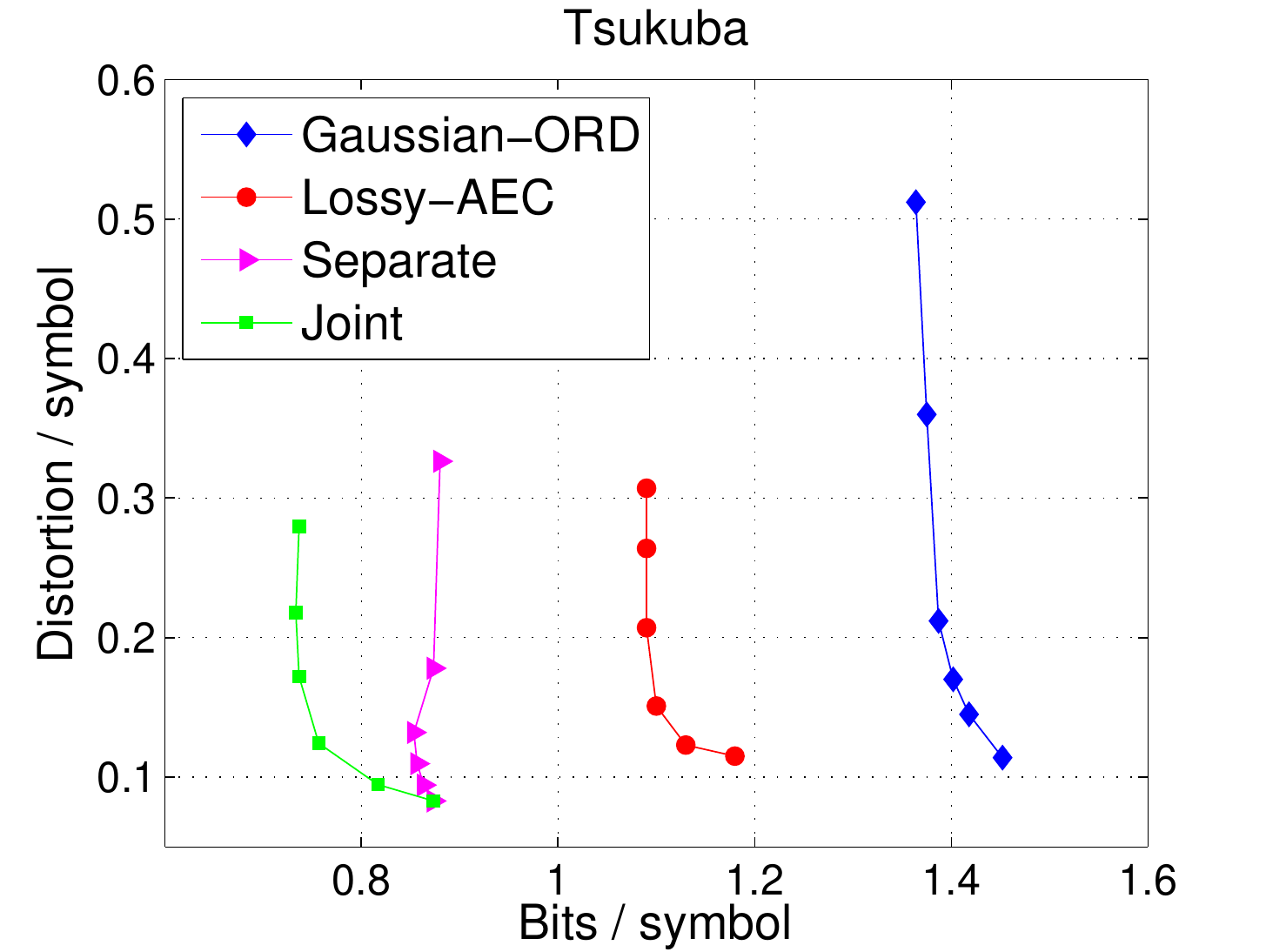}}
\end{minipage}
\hfill
\begin{minipage}[b]{0.48\linewidth}
  \centering
  \centerline{\includegraphics[width=4.5cm]{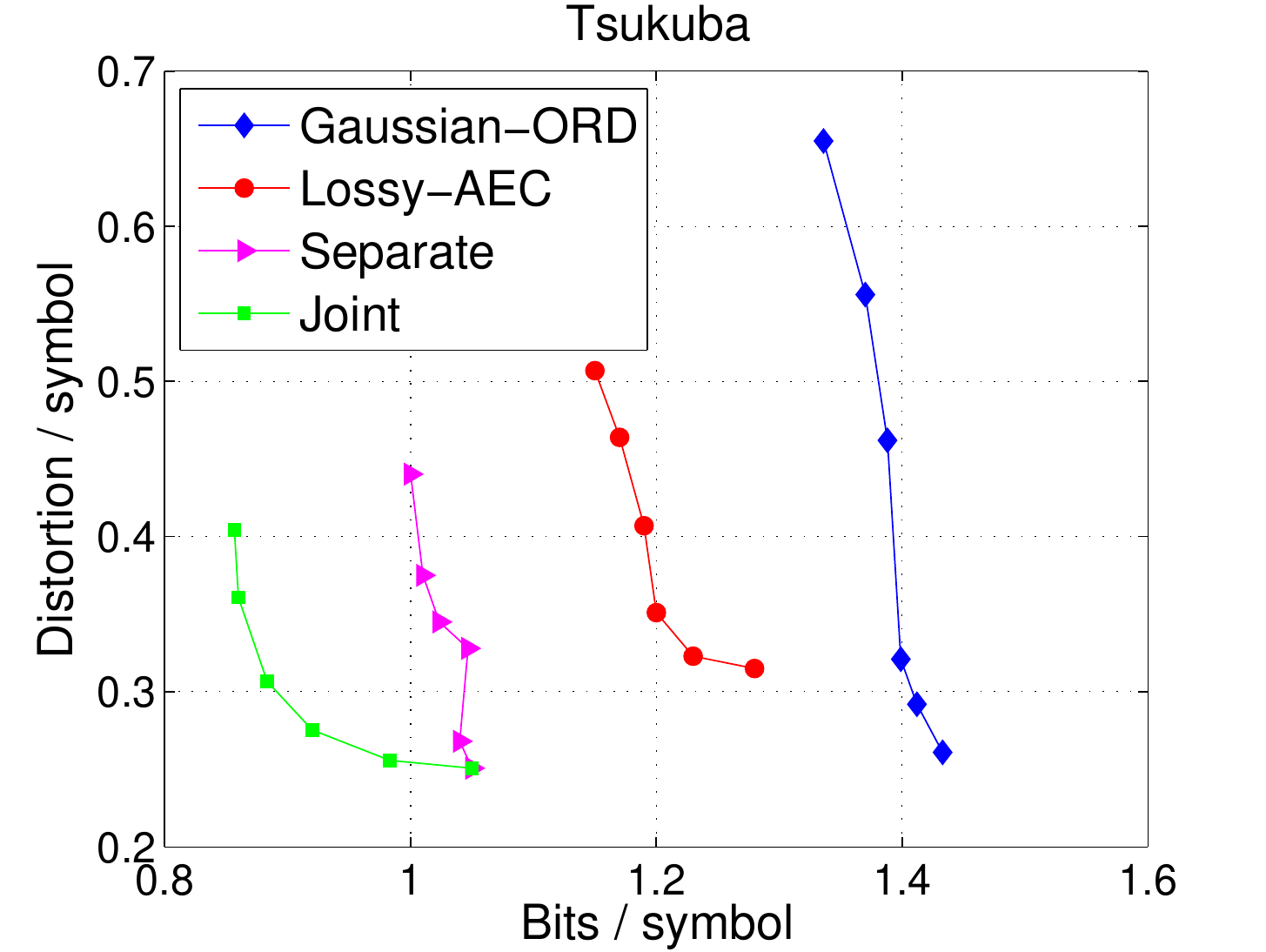}}
\end{minipage}

\caption{Rate distortion curve of \texttt{Dude} and \texttt{Tsukuba}. Left column: noise ratio is 10\%. Right column: noise ratio is 30\%.}
\label{fig:RDcurve_synthetic}
\end{figure}

\begin{figure}[t]

\begin{minipage}[b]{.48\linewidth}
  \centering
  \centerline{\includegraphics[width=4.5cm]{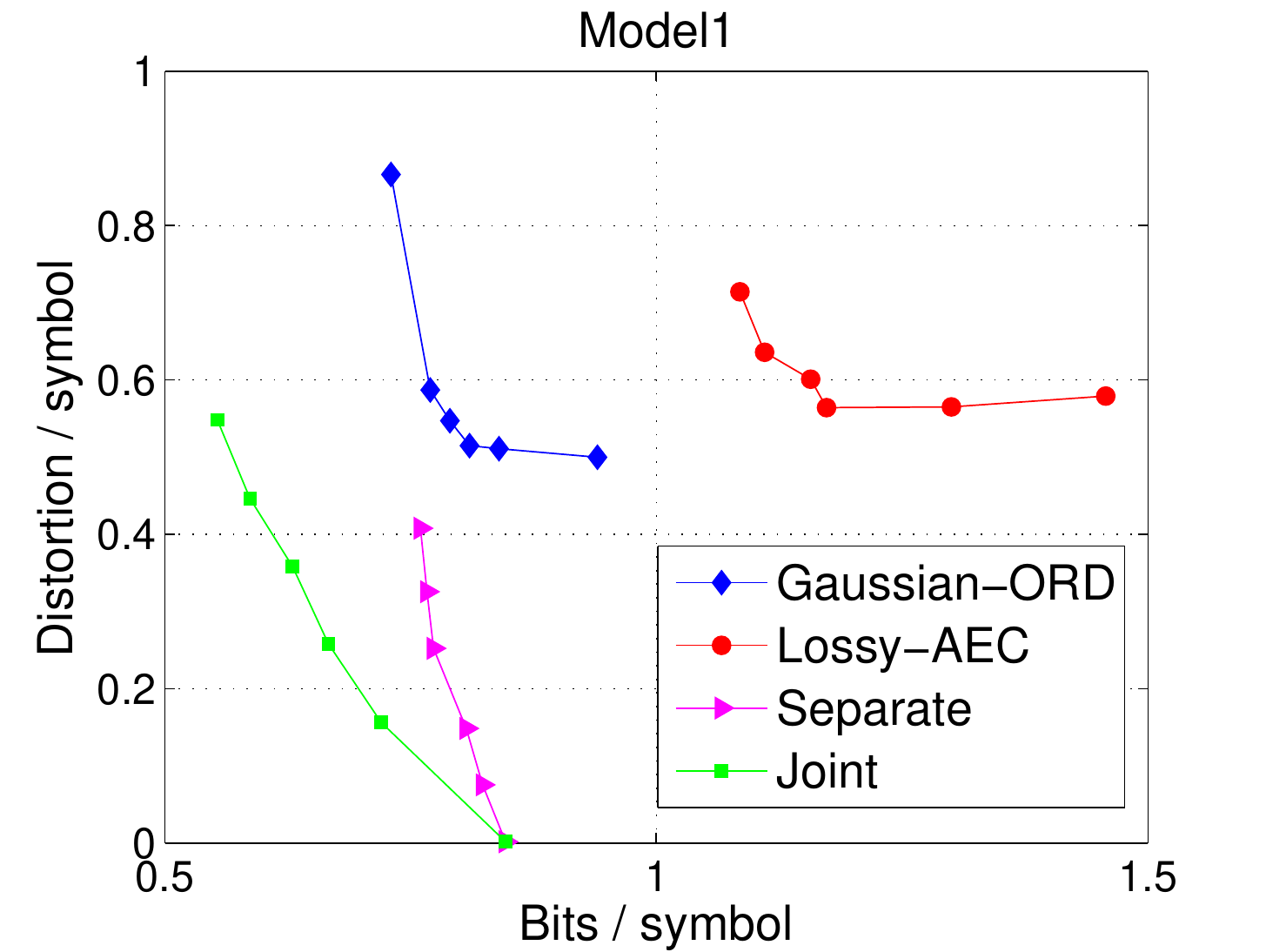}}
\end{minipage}
\hfill
\begin{minipage}[b]{0.48\linewidth}
  \centering
  \centerline{\includegraphics[width=4.5cm]{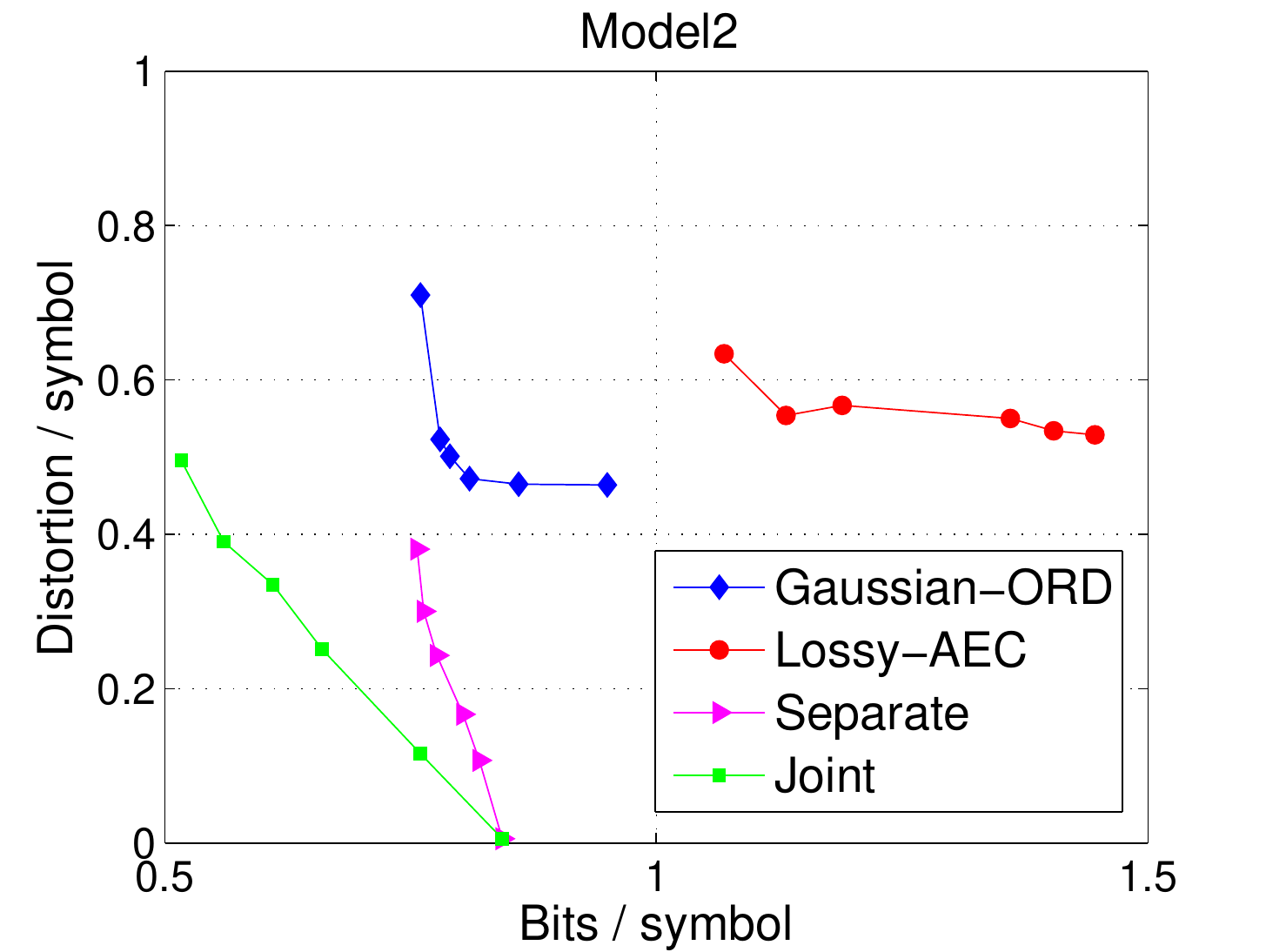}}
\end{minipage}

\vspace{-0.2cm}
\caption{Rate distortion curve of \texttt{Model1} and \texttt{Model2}.}
\label{fig:RDcurve_silhouette}
\end{figure}

We show the rate-distortion (RD) performance of the four comparison schemes in Fig.\;\ref{fig:RDcurve_synthetic} and Fig.\;\ref{fig:RDcurve_silhouette}.
For \texttt{Gaussian-ORD}, RD curve was generated by adjusting the coding parameters of \texttt{ORD} \cite{lai2010arbitrary}. 
For \texttt{Lossy-AEC}, RD-curve was generated by adjusting the strength of contour approximation \cite{yuan2015contour}.
For \texttt{Joint}, RD curve was obtained by varying $\lambda$.
For \texttt{Separate}, we used the same fixed $\beta$ as in \texttt{Joint} to get the best denoising performance (MAP solution), then we increased $\beta$ to further smooth the contour to reduce the bit rate and obtained the RD curve.

We see that \texttt{Joint} achieved the best RD performance for both the depth sequences and the silhouette sequences, demonstrating the merit of our joint approach.
In particular, we save about 14.45\%, 36.2\% and 38.33\% bits on average against \texttt{Separate}, \texttt{Lossy-AEC} and \texttt{Gaussian-ORD} respectively for depth sequences with 10\% noise, and about 15.1\%, 28.27\% and 32.02\% for depth sequences with 30\% noise, and about 18.18\%, 54.17\% and 28.57\% respectively for silhouette sequences. 

Comparing the results with 10\% noise and the results with 30\% noise, we find that more noise generally results in larger bit rate for \texttt{Joint}, \texttt{Separate} and \texttt{Lossy-AEC}.
For \texttt{Gaussian-ORD}, the bit rate does not increase much with more noise, especially the results of \texttt{Tsukuba}.
This was because \texttt{Gaussian-ORD} used a vertex-based contour coding approach; \textit{i.e.}, \texttt{ORD} encoded the locations of some selected contour pixels (vertices).
After denoising using Gaussian filter, the number of selected vertices by \texttt{ORD} are similar for different noise probabilities, resulting in similar bit rates. 

Looking at the results by \texttt{Separate}, when the distortion becomes larger, the bit rate reduces slowly, in some points even increases slightly, especially compared to \texttt{Joint}.
Note that the RD curve of \texttt{Separate} was obtained by increasing $\beta$ to get over-smoothed denoised contour to reduce the bit rate.
It shows that smoother contour dose not always ensure smaller bit rate, which in other hand verifies the importance of the rate term in the problem formulation.
Thus, both the prior term and the rate term are necessary for our contour denoising / compression problem.

Compared to \texttt{Lossy-AEC} which encodes each DCC symbol using a limited number of candidates of conditional probabilities, our proposed \texttt{Joint} and \texttt{Separate} achieve much better coding performance by using PPM with variable length context tree model.
Compared to \texttt{Gaussian-ORD} which encodes the contour by coding some selected vertices, our proposed \texttt{Joint} encodes more contour pixels, but the accurately estimated conditional probabilities enable our contour pixels to be coded much more efficiently than \texttt{Gaussian-ORD}.

\subsection{Performance of Denoising}
\label{subsec:Denoising_performance}

\begin{figure}[t]

\begin{minipage}[b]{.32\linewidth}
  \centering
  \centerline{\includegraphics[width=2.8cm]{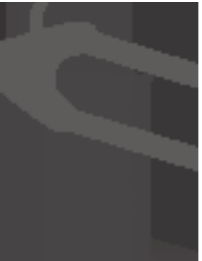}}
  \centerline{(a) Ground truth}\medskip
\end{minipage}
\hfill
\begin{minipage}[b]{0.32\linewidth}
  \centering
  \centerline{\includegraphics[width=2.8cm]{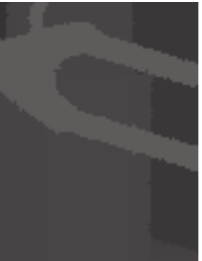}}
  \centerline{(b) 10\% noise}\medskip
\end{minipage}
\hfill
\begin{minipage}[b]{0.32\linewidth}
  \centering
  \centerline{\includegraphics[width=2.8cm]{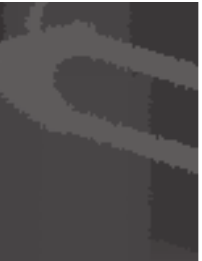}}
  \centerline{(c) 30\% noise}\medskip
\end{minipage}
\vfill

\begin{minipage}[b]{.32\linewidth}
  \centering
  \centerline{\includegraphics[width=2.8cm]{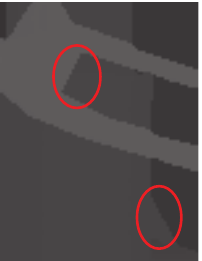}}
  \centerline{(d) \texttt{Gaussian-ORD}}\medskip
\end{minipage}
\hfill
\begin{minipage}[b]{0.32\linewidth}
  \centering
  \centerline{\includegraphics[width=2.8cm]{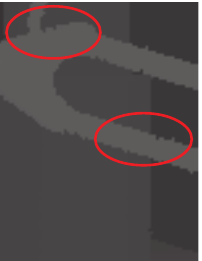}}
  \centerline{(e) \texttt{Lossy-AEC}}\medskip
\end{minipage}
\hfill
\begin{minipage}[b]{0.32\linewidth}
  \centering
  \centerline{\includegraphics[width=2.8cm]{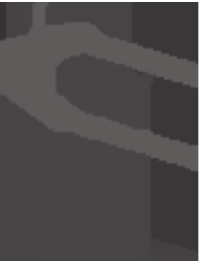}}
  \centerline{(f) \texttt{Joint}}\medskip
\end{minipage}
\vfill

\begin{minipage}[b]{.32\linewidth}
  \centering
  \centerline{\includegraphics[width=2.8cm]{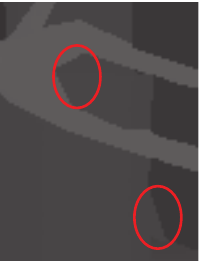}}
  \centerline{(g) \texttt{Gaussian-ORD}}\medskip
\end{minipage}
\hfill
\begin{minipage}[b]{0.32\linewidth}
  \centering
  \centerline{\includegraphics[width=2.8cm]{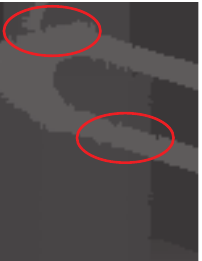}}
  \centerline{(h) \texttt{Lossy-AEC}}\medskip
\end{minipage}
\hfill
\begin{minipage}[b]{0.32\linewidth}
  \centering
  \centerline{\includegraphics[width=2.8cm]{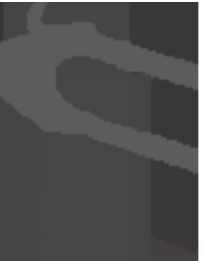}}
  \centerline{(i) \texttt{Joint}}\medskip
\end{minipage}

\caption{Visual denoising results of \texttt{Tsukuba}. Middle row: denoising results with 10\% noise. Last row: denoising results with 30\% noise. (d)\;$\sim$\;(f) at bit rate 1.38, 1.13 and 0.84 bits/symbol respectively. (g)\;$\sim$\;(i) at bit rate 1.39, 1.23 and 0.95 bits/symbol respectively.}
\label{fig:subjective_synthetic}
\end{figure}

\begin{figure}[t]

\begin{minipage}[b]{.48\linewidth}
  \centering
  \centerline{\includegraphics[width=2.8cm]{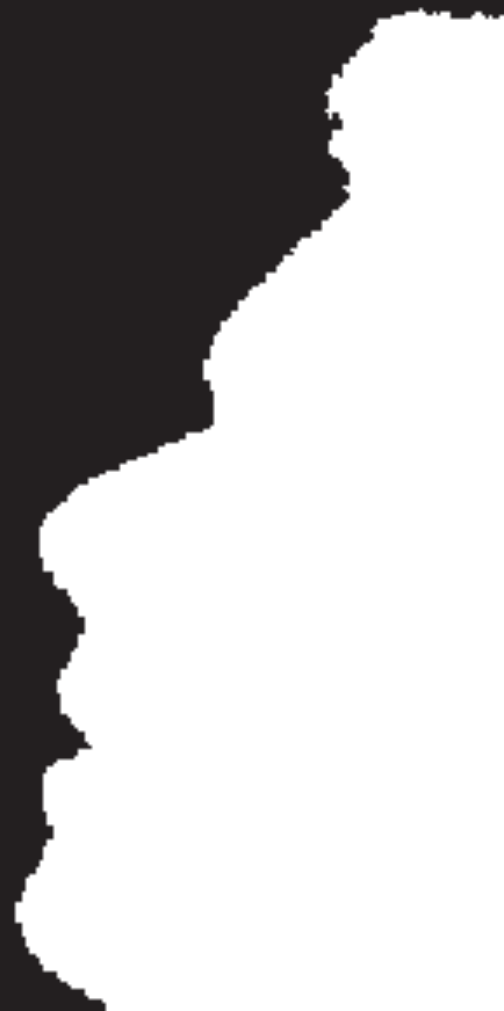}}
  \centerline{(a) Observation}\medskip
\end{minipage}
\hfill
\begin{minipage}[b]{0.48\linewidth}
  \centering
  \centerline{\includegraphics[width=2.8cm]{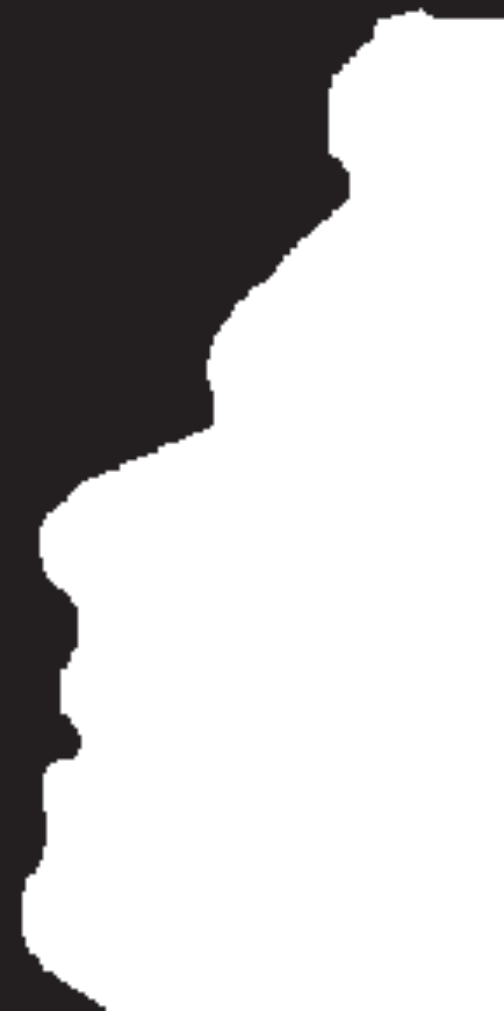}}
  \centerline{(b) MAP solution}\medskip
\end{minipage}
\vfill

\begin{minipage}[b]{.32\linewidth}
  \centering
  \centerline{\includegraphics[width=2.8cm]{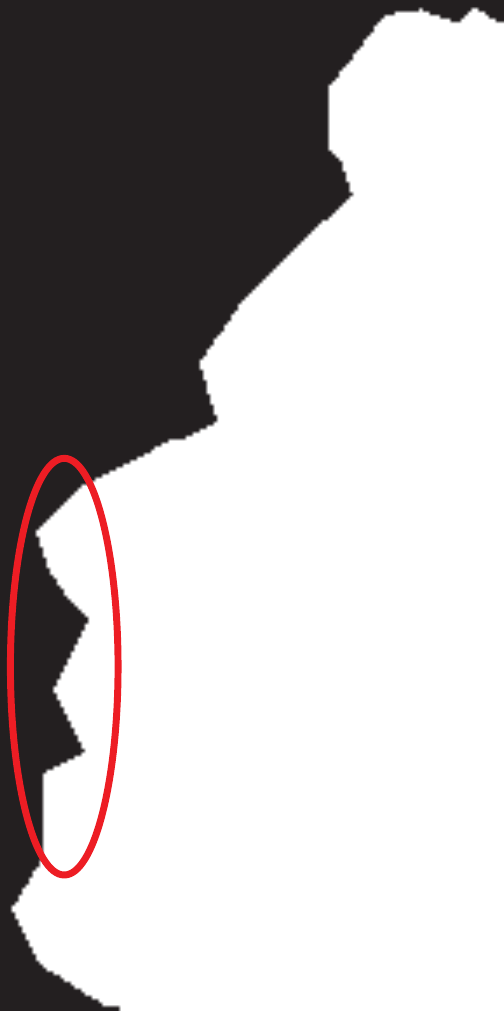}}
  \centerline{(c) \texttt{Gaussian-ORD}}\medskip
\end{minipage}
\hfill
\begin{minipage}[b]{0.32\linewidth}
  \centering
  \centerline{\includegraphics[width=2.8cm]{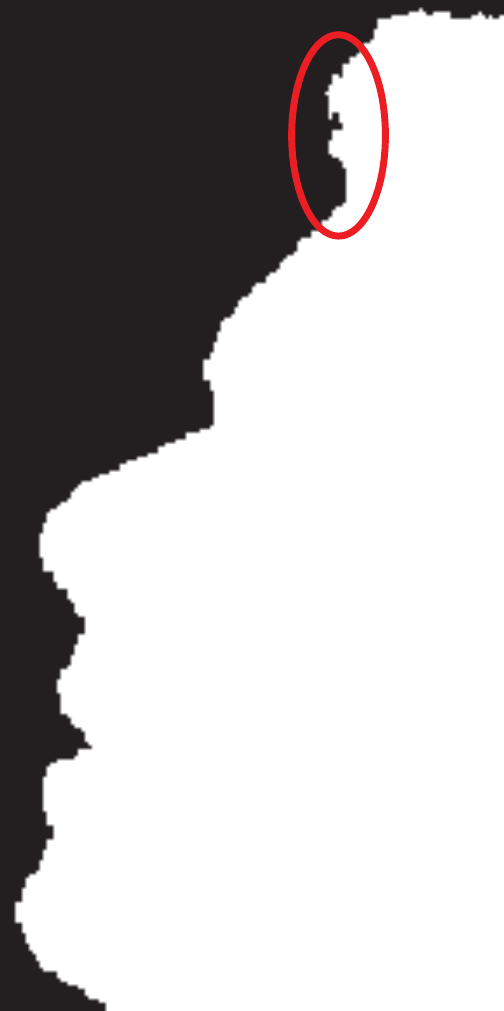}}
  \centerline{(d) \texttt{Lossy-AEC}}\medskip
\end{minipage}
\hfill
\begin{minipage}[b]{0.32\linewidth}
  \centering
  \centerline{\includegraphics[width=2.8cm]{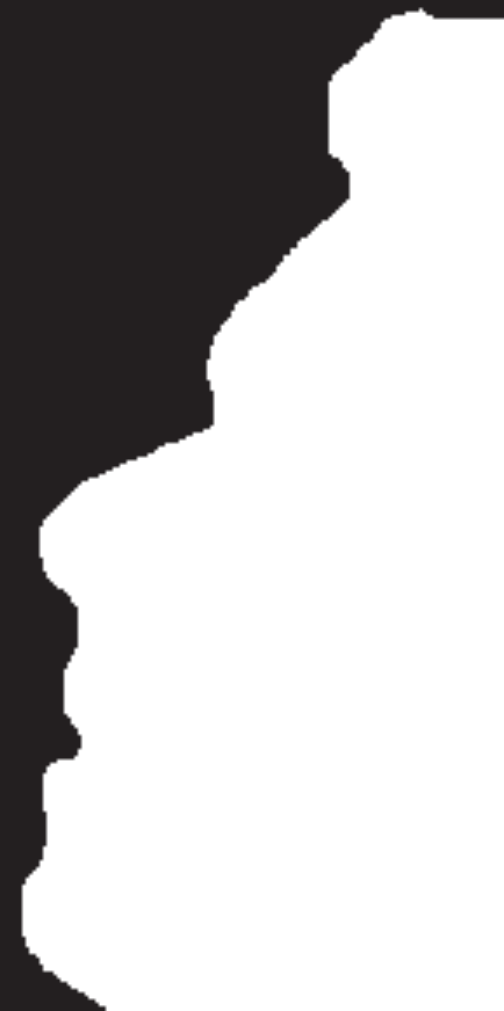}}
  \centerline{(e) \texttt{Joint}}\medskip
\end{minipage}

\caption{Visual denoising results of \texttt{Model1}. (c)\;$\sim$\;(e) at bit rate 0.93, 1.30 and 0.72 bits/symbol respectively.}
\label{fig:subjective_silhouette}
\end{figure}

Fig.\;\ref{fig:subjective_synthetic} and Fig.\;\ref{fig:subjective_silhouette} illustrate the visual denoising results of \texttt{Tsukuba} and \texttt{Model1}.
We see that the denoised contours by \texttt{Joint} are most visually similar to the ground truth contours.
As shown inside the red circle of Fig.\;\ref{fig:subjective_synthetic}(d)\;(g) and Fig.\;\ref{fig:subjective_silhouette}(c), the results by \texttt{Gaussian-ORD} contain lots of staircase shapes.
This was because that the decoded contours were constructed by connecting the selected contour pixels in order, making the result unnatural with too many staircase lines.
\texttt{Lossy-AEC} used a pre-defined irregularity-detection approach to denoise the contours, which failed to remove the undefined noise, \textit{i.e.}, some noise along the diagonal directions of the contours or some noise with high irregularities, as illustrated inside the red circle of Fig.\;\ref{fig:subjective_synthetic}(e)\;(h) and Fig.\;\ref{fig:subjective_silhouette}(d).  

\section{Conclusion}
\label{sec:conclude}
In this paper, we investigate the problem of joint denoising / compression of detected contours in images. 
We show theoretically that in general a joint denoising / compression approach can outperform a separate two-stage approach that first denoises then encodes the denoised contours lossily.
Using a burst error model that models errors in an observed string of directional edges, we formulate a rate-constrained MAP problem to identify an optimal string for lossless encoding.
The optimization is solved optimally using a DP algorithm, sped up by using a total suffix tree (TST) representation of contexts.
Experimental results show that our proposal outperforms a separate scheme noticeably in RD performance.

\bibliographystyle{IEEEbib}
\bibliography{ref}

\end{document}